\documentclass[journal]{IEEEtran}
\usepackage{amssymb}
\usepackage{mathbbold}
\usepackage{mathrsfs}
\usepackage{stmaryrd}
\usepackage{amsfonts}
\usepackage{latexsym}
\usepackage{epsfig}
\usepackage{listings}
\usepackage{booktabs}
\usepackage{subfigure}
\usepackage{amsmath,bm}
\usepackage{picins}
\graphicspath{{figs/}}

\usepackage{algorithm}
\usepackage{algorithmic}
\usepackage{multirow}

\usepackage{amsthm}

\ifCLASSINFOpdf
\else
\fi
\begin{document}
%
\title{Quantitative Analysis of Molecular Transport in the Extracellular Space Using Physics-Informed Neural Network}
%
%
%

\author{Jiayi~Xie,
        Hongfeng~Li,
        Jin~Cheng,
        Qingrui~Cai,
        Hanbo~Tan,
        Lingyun~Zu,
        Xiaobo~Qu,~\IEEEmembership{Member,~IEEE},
        and~Hongbin~Han
\thanks{This work was supported in part by the National Nature Science Foundation of China under Grant 12126601, 12371480, 12171330, 62394310 and 62394311, in part by the R\&D project of Pazhou Lab (Huangpu) under Grant 2023K0608, and in part by the Peking University Medicine Sailing Program for Young Scholars' Scientific \& Technological Innovation under Grant BMU2023YFJHMX013. (Corresponding author: Hongbin~Han)}
\thanks{J.~Y.~Xie is with the Department of Automation, Tsinghua University, Beijing 100084, China and Institute of Medical Technology, Peking University Health Science Center, Beijing 100191, China.}
\thanks{H.~F.~Li and H.~B.~Tan are with the Institute of Medical Technology, Peking University Health Science Center, Beijing 100191, China.}
\thanks{J.~Cheng is with the School of Mathematical Sciences, Fudan University, Shanghai 200433, China.}
\thanks{Q.~R.~Cai and X.~B.~Qu are with the National Integrated Circuit Industry Education Integration Innovation Platform, School of Electronic Science and Engineering (National Model Microelectronics College), Xiamen University, Xiamen 361102, China and Department of Electronic Science, Fujian Provincial Key Laboratory of Plasma and Magnetic Resonance, Xiamen University, Xiamen 361102, China.}
\thanks{L.~Y.~Zu is with the Department of Endocrinology and Metabolism, Department of Cardiology and Institute of Vascular Medicine, Peking University Third Hospital, Beijing 100191, China.}
\thanks{H.~B.~Han is with the Institute of Medical Technology, Peking University Health Science Center, Beijing 100191, China, Department of Radiology, Peking University Third Hospital, Beijing 100191, China, Peking University Third Hospital, Beijing Key Laboratory of Magnetic Resonance Imaging Devices and Technology Beijing 100191, China, and NMPA key Laboratory of Evaluation of Medical Imaging Equipment and Technique, Beijing 100191, China (e-mail: hanhongbin@bjmu.edu.cn)}
\thanks{J.~Y.~Xie and H.~F.~Li contribute equally to this work.}
}
\maketitle

\begin{abstract}
The brain extracellular space (ECS), an irregular, extremely tortuous nanoscale space located between cells or between cells and blood vessels, is crucial for nerve cell survival. It plays a pivotal role in high-level brain functions such as memory, emotion, and sensation. However, the specific form of molecular transport within the ECS remain elusive. To address this challenge, this paper proposes a novel approach to quantitatively analyze the molecular transport within the ECS by solving an inverse problem derived from the advection-diffusion equation (ADE) using a physics-informed neural network (PINN). PINN provides a streamlined solution to the ADE without the need for intricate mathematical formulations or grid settings. Additionally, the optimization of PINN facilitates the automatic computation of the diffusion coefficient governing long-term molecule transport and the velocity of molecules driven by advection. Consequently, the proposed method allows for the quantitative analysis and identification of the specific pattern of molecular transport within the ECS through the calculation of the P{\'e}clet number. Experimental validation on two datasets of magnetic resonance images (MRIs) captured at different time points showcases the effectiveness of the proposed method. Notably, our simulations reveal identical molecular transport patterns between datasets representing rats with tracer injected into the same brain region. These findings highlight the potential of PINN as a promising tool for comprehensively exploring molecular transport within the ECS.
\end{abstract}


%
\IEEEpeerreviewmaketitle

\section{Introduction}
\label{sec:introduction}
\begin{figure*}[!t]
\centerline{\includegraphics[width=1.5\columnwidth]{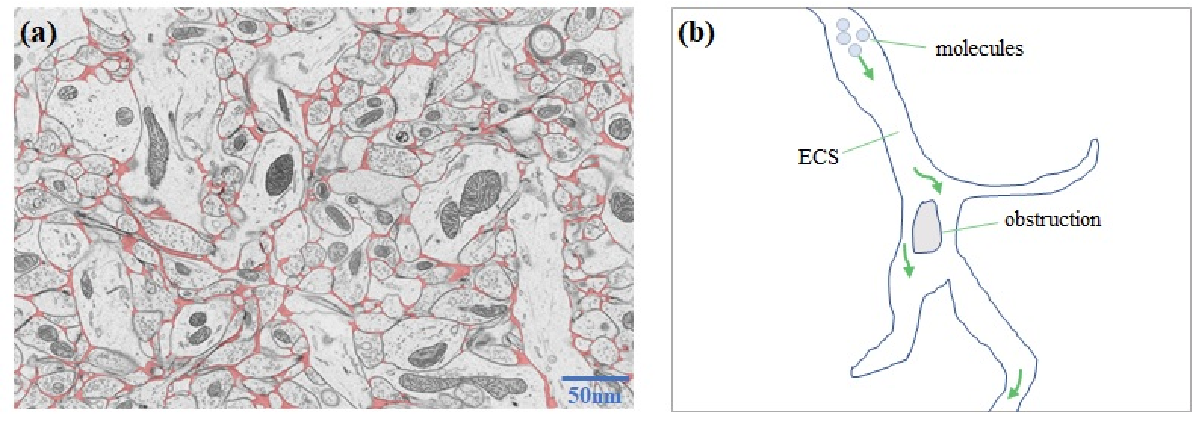}}
\caption{Illustration of the ECS. (a) An electron microscope image of the hippocampal region in the mouse brain. The red area denotes the ECS,  which is an irregular, tortuous narrow space. (b) The schematic diagram of molecular transport within the ECS. The transport of molecules faces various obstructions and is likely driven by both advection and diffusion.}
\label{ecs}
\end{figure*}

%
%
%
%
\IEEEPARstart{T}{he} brain extracellular space (ECS) is an irregular, tortuous nanoscale (about $38$\~{}$64$nm wide) space located between brain cells or between cells and blood vessels (see Fig.~\ref{ecs}(a) for reference), occupying approximately $15$\~{}$20\%$ volume of the living brain. It is the direct space where brain cells rely on to survive~\cite{sykova2008diffusion,wang2021alteration}. Its interior is filled with interstitial fluid (ISF), and the flow of ISF can transport nutrients to cells and take away metabolic products~\cite{iliff2012paravascular}. Current research and understanding of the brain ECS are commonly restricted to limited areas such as the superficial cortex of the brain. However, the structure of the ECS in the vast regions of the deep brain and the laws of molecular transport within the ECS are still areas that have not yet been fully understood~\cite{li2020mechanism,lu2014integrated}.

The transport of molecules within the ECS of the brain has significant influence on high-level brain functional activities such as human memory and emotion~\cite{wiig2012interstitial}. Therefore, quantitative analysis of molecular transport within the ECS has emerged as a research hot topic in this field. The ECS-related research findings offer valuable technical resources and a theoretical foundation for the advancement of new drugs targeting the nervous system, stem cell therapy, drug delivery to the brain interstitium, and physical therapy for the brain interstitium. For example, drug administration through the brain ECS is a new type of encephalopathy treatment method that has gained widespread attention in recent years~\cite{gu2022new,zhou2013protective}. It can overcome the limitations of conventional oral or intravenous administration methods to treat encephalopathy, allowing drugs to bypass the blood-brain barrier and act directly in the cellular microenvironment. In this way, better efficacy can be achieved with lower doses of drugs and reduced risk of systemic toxicity, which brings hope for the development of a variety of central nervous system encephalopathy drugs. There have been some tentative works in this area. Ferguson et al.~\cite{ferguson2007convection} injected drugs into the brain interstitium by applying a positive hydraulic gradient for the clinical treatment of malignant gliomas. Han et al.,~\cite{han2011simple} used a simple diffusion drug administration method to apply a small dose of citicoline in advance through the brain ECS to prevent ischemic neuron damage. In another work, Han et al.,~\cite{gao2022early} independently developed a novel instrument to detect changes in the ultrastructure of the weightless brain, marking a world-first achievement. They elucidated its mechanisms, proposing innovative methods and theories for the protection of astronaut brains.

At present, the prevailing method for detecting the extracellular space of the brain involves utilizing the tracer-based magnetic resonance imaging technology introduced by Han et al. from the Peking University to image pertinent brain areas~\cite{han2014novel}. This method enables the calculation of key parameters such as molecular diffusion rate, tortuosity, and drainage clearance rate of the brain ECS. The quantitative analysis of molecular transport within the ECS involves mathematical modeling applied to relevant brain structure areas. However, the study of molecular transport in the ECS faces challenges due to the lack of accurate mathematical models and efficient optimization algorithms. To address the complex challenges associated with the quantitative analysis of molecular transport within the brain ECS, this paper introduces a novel, efficient, and precise methodology. The molecular transport problem is mathematically formulated using the advection-diffusion equation (ADE), and the resulting inverse problem is effectively tackled through the application of a physics-informed neural network (PINN). Consequently, the specific mode of molecular transport within the ECS can the quantitatively analyzed and identified through the calculation of the P{\'e}clet number. This innovative approach can contribute to a better understanding of the structure and function of the ECS, providing theoretical support and technical advancements for applications in cutting-edge fields such as encephalopathy treatment, new drug research and development, and aerospace medicine.

The contributions of this paper are outlined as follows:
\begin{itemize}
\item[(1)] The problem of molecular transport within the ECS is investigated by formulating and solving an inverse problem derived from the advection-diffusion equation using a physics-informed neural network. The inverse problem can be efficiently solved with the PINN, eliminating the need for complex mathematical formulations and intricate grid settings.
\item[(2)] The diffusion coefficient, governing the long-term transport of molecules, and the velocity of molecules driven by advection movement can be automatically computed through the optimization process of the PINN. Consequently, the proposed method allows for the quantitative analysis and identification of the specific mode of molecular transport within the ECS through the calculation of the P{\'e}clet number.
\item[(3)] Comprehensive experiments are conducted on two datasets and the proposed method demonstrates its capability to accurately derive solutions to the inverse problem. This achievement not only signifies the efficacy of our approach in resolving the advection-diffusion equation but also empowers us to quantitatively compare and analyze distinctive patterns of molecular transport within the ECS.
\end{itemize}

The remainder of the paper is structured as follows: Section \uppercase\expandafter{\romannumeral2} introduces related works, Section \uppercase\expandafter{\romannumeral3} provides a detailed description of the proposed method, Section \uppercase\expandafter{\romannumeral4} presents experiments and result analysis, and in Section \uppercase\expandafter{\romannumeral5}, we draw conclusions and suggest possible future research directions.

\section{Related work}

The brain extracellular space is a complex, narrow and tortuous space between brain cells or between cells and blood vessels. The transport of molecules within the brain ECS faces various obstacles~\cite{vendel2019need}. Fig.~\ref{ecs} illustrates the ECS and the schematic diagram of molecular transport within it. Two primary forms of molecular transport have been identified: advection and diffusion. Advection involves solute movement driven by solvent pressure, independent of molecule size within a certain range~\cite{iliff2019glymphatic}. On the other hand, diffusion is thermally driven, occurring due to a concentration gradient. The rate of diffusion is affected by molecule size, decreasing as the size increases~\cite{sykova2008diffusion}. While some studies suggest that short-distance molecular transport within the ECS is dominated by diffusion and the advection rate of molecules can be negligible~\cite{smith2017test}, others propose that, particularly over longer anatomical distances, advection may play a significant role~\cite{iliff2014impairment}. Han et al.~\cite{wang2019drainage} introduced the tracer-based magnetic resonance imaging method with three-dimensional anisotropic modeling and computing capabilities, revealing regional inconsistencies in molecule transport, indicating coexistence of diffusion and advection in different brain regions. Although molecular transport in the brain ECS is likely driven by both advection and diffusion, the precise quantification and analysis of their relative contributions remain elusive. Accurate quantitative analysis of diffusion and convective motion in distinct brain regions holds promise for precise regulation of material transport, facilitating advancements in clinical applications such as drug delivery through the ECS.

There have been some works that used MRI images to investigate the molecular transport in various brain regions~\cite{elkin2018glymphvis,koundal2020optimal}. These works employ appropriate mathematical models to describe the law of molecular transport in specific brain areas and devise corresponding optimization algorithms for solving these models. Currently, two main types of models are employed to characterize the dynamics of molecular transport in the brain: the advection-diffusion equation and the optimal mass transport model.

The advection-diffusion equation (ADE), sometimes also referred to as the advection-diffusion equation, describes how a molecule transports in a given velocity field, such as the extracellular space~\cite{liu2007stability}. The ADE can be expressed as the following formula:
\begin{equation}
\frac{\partial C(x,t)}{\partial t}+v\cdot\nabla C(x,t)=D\triangle C(x,t), x\in\Omega, t\geq0\label{eq1}
\end{equation}
where $t\in\left[0,1\right]$ is the time, $x$ is the location, $C\left(x,t\right):\mathrm{\Omega}\times\left[0,1\right]\rightarrow\mathbb{R}$ is the molecular concentration in the region $\mathrm{\Omega}\subseteq\mathbb{R}^3$, $v\left(x,t\right):\mathrm{\Omega}\times \left[0,1\right]\rightarrow\mathbb{R}$ is the moving advection velocity of the molecule, and $D$ is the diffusion coefficient. Utilizing MRI image data obtained through the magnetic tracing method, Han et al.,\cite{wang2019stimulation} computed the diffusion coefficient $D$ using model~(\ref{eq1}) without incorporating the advection term. However, the accuracy of the obtained diffusion coefficient $D$ is high in brain areas dominated by molecular diffusion motion but less precise in other regions. This approach neglects the simultaneous consideration of both advection and diffusion forms of motion, limiting its qualitative analysis of molecular transport in the ECS. Therefore, enhancing this method by incorporating factors such as advection velocity has the potential to significantly improve its accuracy in modeling molecular transport within the ECS.

Optimal mass transport (OMT) addresses the task of transporting a mass distribution from one configuration to another by minimizing a specified cost function. The OMT problem was initially formulated by Gaspard Monge in 1781~\cite{monge1781memoire}, and later, in 2000, Benamou and Brenier reformulated OMT, integrating it into the framework of computational fluid dynamics (CFD)~\cite{benamou2000computational}. The optimal mass transport model can be expressed as the following optimization problem:
\begin{align}
&\inf_{C,v} \int^{1}_{0}\int_{\Omega}C(x,t)\parallel v(x,t)\parallel^{2}dxdt \label{eq3} \\
&\text{s.t.,}\frac{\partial C(x,t)}{\partial t}+v\cdot\nabla C(x,t)=D\triangle C(x,t) \label{eq4}
\end{align}

By solving the above model, the molecular concentration $C\left(t_i,x\right)$ and the advection velocity $v\left(t_i,x\right)$ in the brain area at any time $t_i$ can be calculated. Based on this model, Ratner, Elkin and others conducted a series of studies on molecular transport rules in the lymphoid system~\cite{fan2020effect,ratner2017cerebrospinal,ratner2015optimal,chen2023visualizing}. After solving the model, the molecular concentration $C\left(t,x\right)$ and velocity $v\left(t,x\right)$ can be used to make predictions and 3D visualize changes in molecular concentration and velocity. However, Ratner et al.~\cite{ratner2017cerebrospinal,ratner2015optimal} only considered the influence of diffusion or advection terms on molecular transport in their model. Though Chen et al.~\cite{chen2023visualizing,chen2022cerebral}, considered both diffusion and advection terms in their work, the diffusion coefficient $D$ is manually set as a constant. As a result, only information related to advection can be obtained.

The models discussed above are typically addressed using traditional numerical methods, such as the finite element method~\cite{mardal2022mathematical}. These methods necessitate meticulous meshing of the region of interest (ROI) and the formulation of assumptions regarding boundary conditions. Notably, meshing in brain modeling poses specific challenges due to the intricate and irregular nature of the brain's geometry~\cite{valnes2020apparent}. Recent studies have embraced the use of deep neural networks~\cite{li2018neural}, particularly physics-informed neural networks (PINNs), which have gained popularity in recent years for solving partial differential equations (PDEs). The concept of PINNs is first proposed by Raissi et al.\cite{raissi2019physics}, wherein the PDE describing the problem is integrated into the loss function of a neural network. This additional constraint compels the neural network to progressively approximate a solution that adheres to the underlying law of the PDE~\cite{raissi2018hidden}. PINNs can derive solutions solely based on the PDE and the provided initial and boundary conditions, eliminating the need for additional measurement data. Additionally, with physics-based synthetic data, PINNs can be trained effectively even in scenarios where real data is scarce or unavailable, addressing the challenge posed by the limited availability of large datasets~\cite{yang2023physics,cai2023bloch}. Compared with traditional numerical methods, PINNs offer several advantages: 1) They avoid the complexities of seeking mathematical solutions and data grid planning, relying solely on the backpropagation algorithm for optimizing the network to achieve high-precision numerical solutions; 2) They enable rapid inference; 3) They demonstrate potential advantages in handling high-dimensional data; 4) The same processing method can be used for solving both forward and inverse PDE-related problems, simplifying the overall process~\cite{kapoor2023physics,cui2023knowledge}.

\begin{figure*}[!t]
\centerline{\includegraphics[width=1.5\columnwidth]{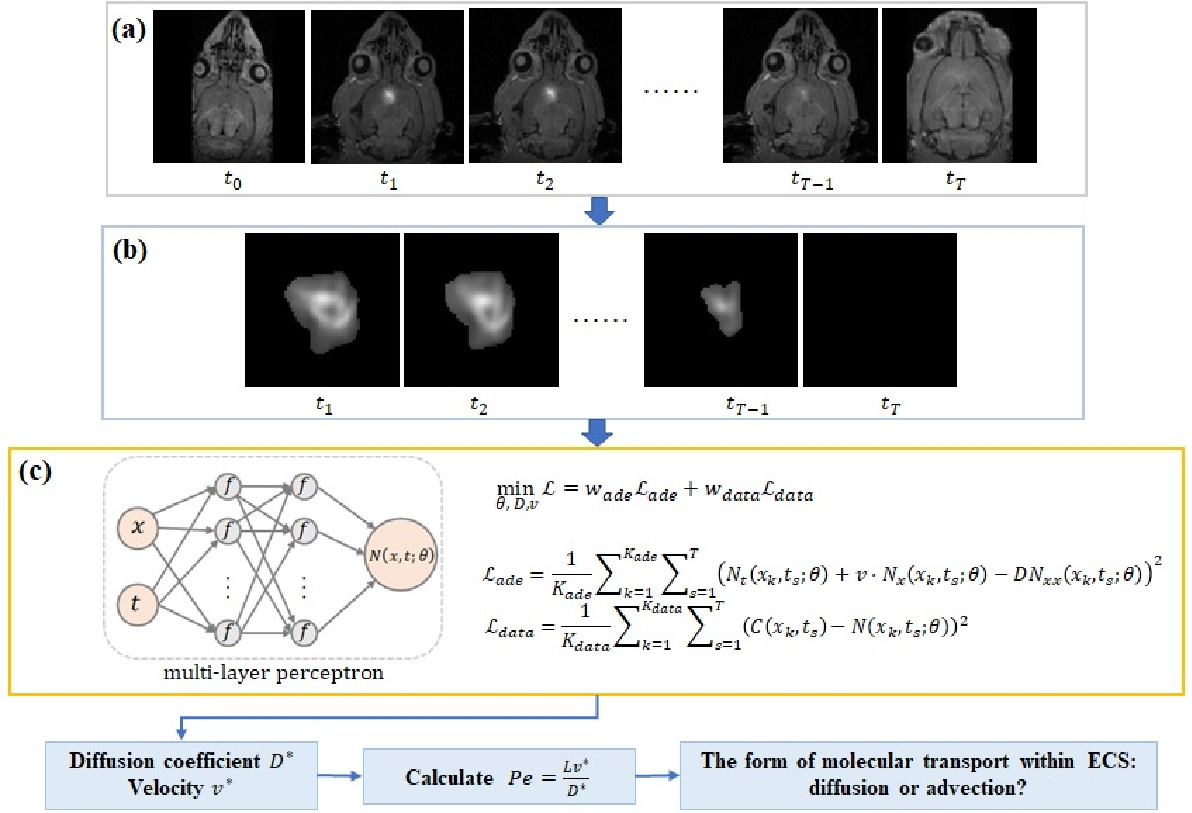}}
\caption{The framework of the proposed method. The initial step in our methodology involves acquiring MRI data at various time points $t_0, t_1, t_2, \cdots, t_{T-1}, t_T$. Subsequently, we perform image preprocessing to eliminate unnecessary regions and align images through displacement registration. Using the preprocessed data, we develop a physics-informed neural network designed to solve the inverse problem arising from the advection-diffusion equation. The primary function of the network is to automatically determine the diffusion coefficient $D$ and velocity $v$. As a result, we can calculate the P{\'e}clet number, identifying the specific mode of molecular transport within the ECS. (a) MRI data obtained at different time points $t_0, t_1, t_2, \cdots, t_{T-1}, t_T$. (b) MRI data after image preprocessing. (c) A physics-informed neural network.}
\label{framework}
\end{figure*}

Several works have utilized PINNs to model the fluid mechanics in brain glymphatic system or blood vessels. In the study conducted by Zapf et al.~\cite{zapf2022investigating}, the authors aimed to estimate the diffusion coefficient related to the long-distance movement of molecules in the human brain. They employed a PINN, which was constrained by model (\ref{eq1}) without the advection term, utilizing MRI data. Notably, the experimental results demonstrated a remarkable proximity between the outcomes obtained through PINN and those derived from the finite element method. This close alignment strongly attests to the effectiveness of the PINN approach in modeling and estimating the molecular transport within the human brain. Hertena et al.~\cite{van2022physics} applied a PINN to quantitatively analyze myocardial perfusion with MRI data, estimating relevant kinetic parameters. PINN is constrained by a pair of coupled ordinary differential equations that describe the evolution of contrast agent concentration over time. Experiments have shown that this method can obtain results consistent with clinical practice. Sarabian et al.~\cite{sarabian2022physics} proposed a physics-informed deep learning framework that combines sparse clinical measurements with one-dimensional (1D) reduced-order model (ROM) simulations to generate high-resolution, physically consistent brain hemodynamic parameters. The approach demonstrated the ability to estimate subject-specific cerebral hemodynamic variables with high accuracy, even in the absence of knowledge about inlet and outlet boundary conditions. Rosen et al.~\cite{oszkinat2022uncertainty} developed an approach based on PINNs to estimate a blood alcohol signal from a transdermal alcohol signal. The proposed method is able to estimate the blood alcohol signal and quantify the uncertainty in the form of conservative error bands. Zhang et al.~\cite{zhang2023physics} tailored a physics-informed neural network using the continuity equation and Navier-Stokes equations as governing equations for 4D hemodynamics prediction. This deep learning framework facilitates real-time, precise, and resilient mapping from vessel structure and time series to the development of velocity and pressure fields, allowing for the comprehensive visualization of 4D blood flow dynamics under the influence of various input parameters. The aforementioned studies that utilized PINN were centered around the human brain glymphatic system or blood flow, which differs significantly from the molecular transport in the brain ECS. Consequently, there is a need for the construction and optimization of a PINN specifically tailored to address the unique aspects of molecular transport within the brain ECS. In this study, we investigate the molecular transport within the ECS using temporally sparse T1-weighted dynamic contrast enhanced MRI data and PINN. The PINN is constrained by the advection-diffusion equation, encompassing both advection and diffusion terms, distinguishing it from previous works that employed PINN with MRI data.

In summary, despite existing studies on the rules of molecular transport in human brain or blood vessels, challenges such as incomplete mathematical models for simulating molecular transport within the ECS, complex optimization algorithms, and high-demand computing requirements persist. To advance our understanding and address these challenges, it is imperative to delve further into the intricacies of molecular transport in the brain ECS, leveraging the latest research progress in this field. This paper aims to address the challenges associated with the quantitative analysis of molecular transport within the brain ECS. The study formulates the problem of molecular transport through the mathematical lens of the advection-diffusion equation. Subsequently, a PINN is employed to efficiently solve the resulting inverse problem, enabling a rigorous quantitative analysis of molecular transport within the ECS.

\begin{figure*}[!http]
\centerline{\includegraphics[width=1.5\columnwidth]{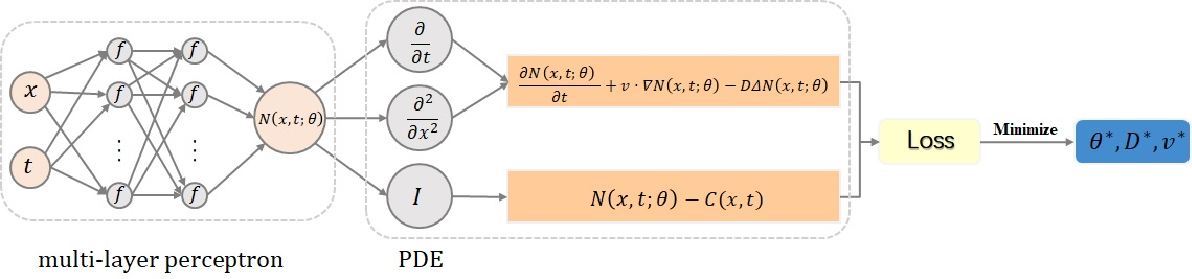}}
\caption{The framework of PINN for solving the problem~(\ref{eq1-2}) and~(\ref{eq2-2}). A multi-layer perceptron (MLP) is constructed to approximate the solution C(x,t) in (\ref{eq1-2}), while optimizing the network by minimizing a loss function that incorporates the advection-diffusion equation as a constraint.}
\label{pinn}
\end{figure*}

\section{Method}

In this section, we elaborate on the proposed method for modeling and quantitatively analyzing molecular transport within the brain ECS, as depicted in Fig.~\ref{framework}. The initial step of our approach involves obtaining MRI data at various time points $t_0, t_1, t_2, \cdots, t_{T-1}, t_T$. Subsequently, we perform image preprocessing to eliminate unnecessary regions and align images through displacement registration. Using the preprocessed data, we develop a physics-informed neural network designed to solve the inverse problem arising from the advection-diffusion equation. The primary function of the network is to automatically determine the diffusion coefficient $D$ and velocity $v$. Consequently, we can calculate the P{\'e}clet number, identifying the specific mode of molecular transport within the ECS.

\subsection{Problem Statement}

As discussed earlier, current research on analyzing the law of molecular transport within the ECS is characterized by a lack of precise mathematical models and corresponding high-efficiency optimization algorithms. On one hand, certain works~\cite{wang2019stimulation,ratner2017cerebrospinal,ratner2015optimal} have approached the problem with mathematical models that only account for the influence of either diffusion or advection terms on molecular transport. These approaches, however, contradict the observed phenomenon that the transport of molecules in the brain ECS is driven by both advection and diffusion. On the other hand, some works~\cite{chen2023visualizing} considered both diffusion and advection terms; however, a notable limitation exists in that the diffusion coefficient is manually set as a constant. Consequently, obtaining precise information about the diffusion process becomes challenging, contributing to a gap in our understanding of molecular transport dynamics within the ECS. Notably, with the exception of the method proposed in~\cite{wang2019stimulation}, the aforementioned methods were not specifically designed for analyzing molecular transport within the ECS.

To resolve the above problem, we propose to describe the molecular transport within ECS with the following equation:
\begin{align}
&\frac{\partial C(x,t)}{\partial t}+v\cdot\nabla C(x,t)=D\triangle C(x,t)\label{eq1-2} \\
&\text{s.t.,}\ C(x,t_0)=C_0(x),\ C(x, t_{T})=C_{T}(x) \label{eq2-2}\\
&\text{\ \ \ \ \ for}\ (x,t)\in \Omega\times[0,T]\nonumber
\end{align}
where $t\in[0,T]$ is the measurement time, $T$ is the end time, $\Omega\in \mathbb{R}^3$ is the predefined region of interest (ROI) (i.e., brain ECS here) and $x$ denotes any location within $\Omega$. $C(x,t): \Omega\times[0,T]\rightarrow\mathbb{R}$ and $v\in\mathbb{R}$ are the time-dependent molecular concentration and velocity, respectively, and $D>0$ is the diffusion coefficient. (\ref{eq1-2}) represents the advection-diffusion equation (ADE) commonly used in fluid dynamics, encompassing both advection and diffusion motions. The accompanying initial conditions, denoted by (\ref{eq2-2}), are only partially known and govern the behavior of the ADE.

The data used for stimulating the molecular transport within the ECS are T1-weighted dynamic contrast enhanced MRI (DCE-MRI) data collected with the tracer-based magnetic resonance imaging technology~\cite{han2014novel} performed on two rats. We obtained a series of MRI data of rats measured at different time points $t_{i}$. It is the usual practice to assume that the molecular concentration $C(x,t)$ is proportion to the intensity of pixels in MRI~\cite{chen2023visualizing}, and thus we directly take the image intensity as the concentration of molecules.

The major difference between our work and previous ones is that the diffusion coefficient $D$ and velocity $v$ are automatically calculated along with the optimization of a neural network instead of manually provided, which is expected to be more precise for quantitatively analyzing the molecular transport. To efficiently solve the model (\ref{eq1-2}) and (\ref{eq2-2}), PINN is employed to solve the resulting inverse problem so as to obtain the key parameters, i.e., $D$ and $v$, for describing the molecular transport within the ECS.

\subsection{Physics-Informed Neural Networks}

Developed in recent years, physics-informed neural networks (PINNs) offer a novel approach to solving partial differential equations (PDEs) by integrating them into the optimization process of neural networks. This integration imposes an additional constraint, compelling the neural networks to progressively approximate solutions in accordance with the underlying laws of the PDEs~\cite{raissi2018hidden}. PINNs excel in obtaining solutions based solely on the PDEs and provided initial and boundary conditions, eliminating the need for additional measurement data. The challenge posed by the problem in (\ref{eq1-2}) and (\ref{eq2-2}) involves determining the solution for $C(x,t)$, $D$, and $v$, constituting a nonlinear ill-posed inverse problem. This complexity makes it challenging for traditional numerical solvers, such as the finite element method~\cite{mardal2022mathematical}.

Following the PINN framework, we seek to approximate the solution $C(x,t)$ of the advection-diffusion equation in (\ref{eq1-2}) using a deep neural network. The architecture of the PINN constructed in this paper is visualized in Fig.~\ref{pinn}.  Previous studies have demonstrated the effectiveness of a multi-layer perceptron (MLP) in approximating solutions to PDEs~\cite{raissi2019physics,moser2023modeling}. Accordingly, we construct a MLP network with $4$ hidden layers, each containing 32 nodes. We denote the network as $N(x,t;\theta)$, expressed as follows:
\begin{equation}\label{eq3-0}
\begin{split}
&N(x,t;\theta)=f(W_L(f(W_{L-1}(\cdots f(W_0([x;t])+b_0))\\
&\ \ \ \ \ \ \ \ \ \ +b_{L-1}))+b_L)
\end{split}
\end{equation}
where $f$ is an element-wise nonlinear activation function, $\theta=\{W_l,b_l\}^{L}_{l=1}$ represents the network parameters, encompassing weights $W_l$ and biases $b_l$ ($l=0,1,\cdots,L$) that need optimization. $L$ is set to $4$ here.

The input for the network is the location $x_i$ at different time points $t_i$ ($i=0,1,\cdots,T$) and the output is the approximation of molecular concentration $C(x_i,t_i)$, diffusion coefficient $D$ and velocity $v$. The loss function of the network consists of two items: $\mathcal{L}_{ade}$ for computing the residual of the ADE and $\mathcal{L}_{data}$ for calculating the fidelity between the predictions and the limited measurement data, which can be defined as follows:
\begin{equation}
\mathcal{L}=w_{ade}\mathcal{L}_{ade}+w_{data}\mathcal{L}_{data}\label{eq3}
\end{equation}
where $w_{ade}$ and $w_{data}$ are weights for the three items, respectively, and $\mathcal{L}_{ade}$ and $\mathcal{L}_{data}$ are defined as below:
\begin{align}\label{eq4}
\begin{array}{ll}
     \mathcal{L}_{ade}=\frac{1}{K_{ade}}\sum^{K_{ade}}_{k=1}\sum^{T}_{s=1} (N_t(x_k,t_s;\theta)+v\cdot N_x(x_k,t_s;\theta)\\
     \ \ \ \ \ \ \ \ \ \ -DN_{xx}(x_k,t_s;\theta))^2\\
     \mathcal{L}_{data}=\frac{1}{K_{data}}\sum^{K_{data}}_{k=1}\sum^{T}_{s=1}(C(x_k,t_s)-N(x_k,t_s;\theta))^2
     \end{array}
\end{align}
The first term in (\ref{eq4}) represents the mean squared residual of the ADE, calculated at $K_{ade}$ randomly selected points located at $x_k$ across all time within the ROI. The second in (\ref{eq4}) quantifies the fidelity between the limited $K_{data}$ measured data and the predictions of the network $N(x_k,t;\theta)$ over all time. Training a PINN involves minimizing the total loss in (\ref{eq3}) using optimization algorithms, such as Adam~\cite{kingma2014adam}, to obtain optimal parameters $\theta$ and solutions for the coupled problem in~(\ref{eq1-2}) and~(\ref{eq2-2}). To facilitate the automatic learning of the diffusion coefficient $D$ and velocity $v$, we incorporate these two variables as additional parameters to be optimized within the network. This approach allows for the automatic and precise estimation of $D$ and $v$ during the network training process.

\section{Experimental results and analysis}

In this section, we evaluate the efficacy of the proposed method using two datasets comprising T1-weighted DCE-MRI data collected from rats. The availability of two distinct MRI datasets allows for a comparative analysis of potential variations in molecular transport within the ECS. The implementation of PINN is carried out in PyTorch~\cite{paszke2019pytorch}. All experiments are conducted on a desktop equipped with an Intel i7-10700 CPU and 32 GB RAM. We begin by providing insights into the data acquisition process and detailing the image preprocessing method. Subsequently, we present and analyze the experimental results obtained from the two datasets.

\subsection{Data Acquiring}

Clinical experiments were carried out based on the experimental equipment of Peking University Health Science Center, and MRI scans were performed on rats to collect the MR image data. The detailed experimental process is described below.

\begin{figure}[!t]
\centerline{\includegraphics[width=\columnwidth]{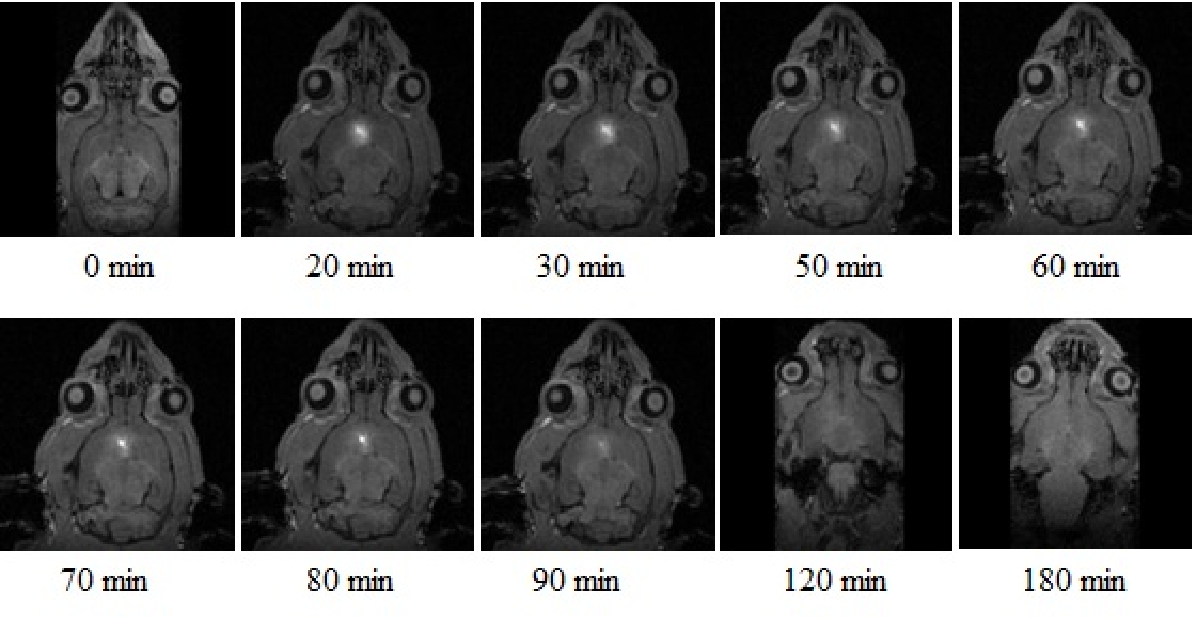}}
\caption{Example MRI (coronal plane) obtained at different time points (i.e., $t=0, 20, 30, 50, 60, 70, 80, 90, 120, 180$ mins) from the dataset-1. Note that at $t=0$ min, the image was acquired before the tracer injection.}\label{mri-N5-20}
\end{figure}

A $3.0$T MRI system (GE750, General Electric, U.S.) equipped with a small animal coil was utilized with magnetization-prepared rapid acquisition gradient echo sequences (MPRAGE) for image acquisition. The parameters were configured as follows: a repetition time of $1,500$ ms, a flip angle of $9^{\circ}$, a field of view measuring $30$ mm, a slice thickness of $0.5$ mm, a resolution of $512\times96$, and a voxel size of $0.5\times0.5$ mm.

The research adhered to national guidelines and received approval from the Ethics Committee of Peking University Health Science Center (Approval No. PUIRB-LA2023256). The experiments involved male Sprague Dawley rats weighing between 280-360 g. Anesthesia was induced in rats through intraperitoneal injection of chloral hydrate (400 mg/kg), followed by fixation in a stereotactic coordinate system (Lab Standard Stereotaxic-Single, Stoelting Co, Illinois, USA). Prior to injection, MRI scans were conducted to confirm the puncture position and obtain a baseline reference image.

A 2 $\mu$l dose of Gd-DTPA was gradually injected into the caudate nucleus of the brain at a rate of 0.2 $\mu$l/min, guided by pre-scan images. According to the previous standard scanning settings of our laboratory, the total scanning time span was set to 4 hours, with the first scan occurring 15 minutes after tracer injection, followed by scans every 15 minutes in the first hour, and every 30 minutes in the subsequent 2-3 hours. Subsequently, scans were conducted every 4 hours. Real-time monitoring of the rat's status during scanning ensured continuous anesthesia. Thus, MRI scans were performed at various time points post-injection.

\begin{figure}[!t]
\centerline{\includegraphics[width=\columnwidth]{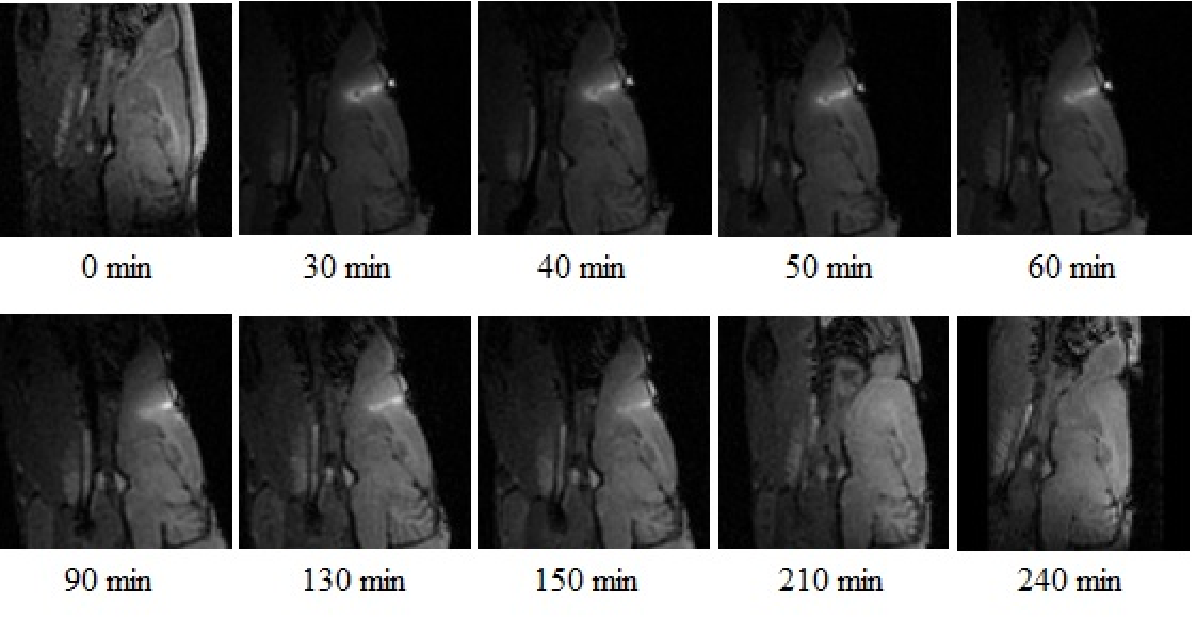}}
\caption{Example MRI (coronal plane) obtained at different time points (i.e., $t=0, 30, 40, 50, 60, 90, 130, 150, 210, 240$ mins. Note that at $t=0$ min, the image was acquired before the tracer injection.) from the dataset-2.}\label{mri-N8-30}
\end{figure}

With the aforementioned clinical protocol, two datasets were generated, each containing T1-weighted DCE-MRI data of a rat with Gd-DTPA injected into the region of caudate nucleus. For convenience, these datasets are denoted as dataset-1 and dataset-2, respectively. The dimensions of each data in both datasets are $512\times512\times52$. Exemplary images from both datasets are presented in Fig.~\ref{mri-N5-20} and~\ref{mri-N8-30} for reference.

\subsection{Image Preprocessing}

Before utilizing the acquired MRI data to assess the proposed method, image preprocessing steps were applied to exclude unnecessary regions and register images with displacement. Specifically, post-injection images from MRI scans were compared with baseline images after grayscale calibration, mutual information-based image registration, and histogram equalization. All post-injection images underwent automatic rigid transformation, similarity measurement, high-order interpolation, and adaptive stochastic gradient descent optimization. The resulting images were then subtracted from the pre-scanned image.

\begin{figure}[!t]
\centerline{\includegraphics[width=\columnwidth]{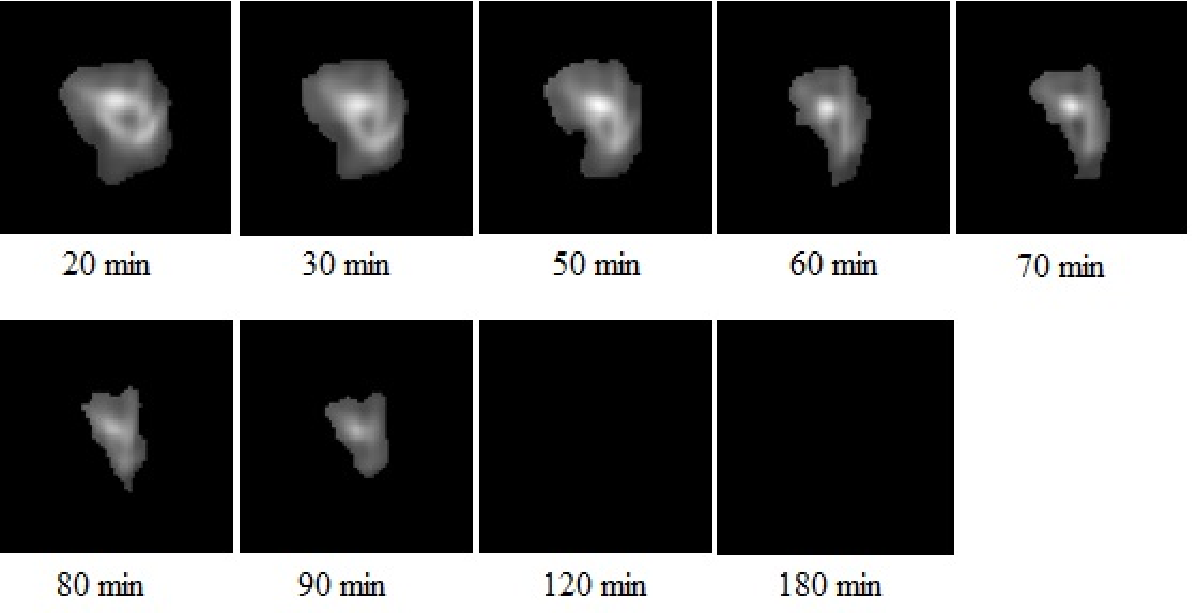}}
\caption{Example MRI data (coronal plane) at different time points (i.e., $t=20, 30, 50, 60, 70, 80, 90, 120, 180$ mins) after image preprocessing from the dataset-1.}\label{mri-N5-20-seg}
\end{figure}

\begin{figure}[!t]
\centerline{\includegraphics[width=\columnwidth]{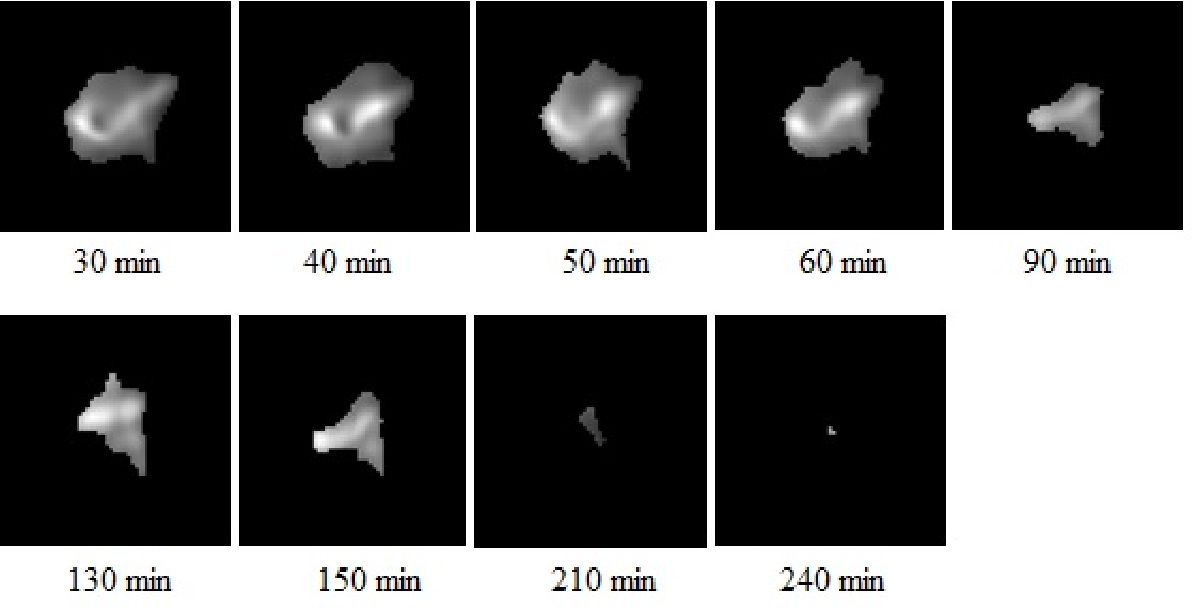}}
\caption{Example MRI data (coronal plane) at different time points (i.e., $t=30, 40, 50, 60, 90, 130, 150, 210, 240$ mins) after image preprocessing from the dataset-2.}\label{mri-N8-30-seg}
\end{figure}

The ``bright area" identified by establishing a seed point and a threshold in the area of interest, was considered associated with the tracer. New sets of post-processed MR images were generated in the horizontal, sagittal, and coronal planes, each with a slice thickness of 1 mm. The image processing steps were implemented using MATLAB-based software developed by ourself. For a more detailed description of data preprocessing, please refer to our previous work~\cite{han2014novel}. Exemplary MRI data at different time points after image preprocessing from the two datasets are displayed in Fig.~\ref{mri-N5-20-seg} and~\ref{mri-N8-30-seg}, respectively.

\subsection{Results on the Dataset-1}

On the dataset-1, the network parameters (i.e., weights and biases) are initialized using the Glorot initialization method and the $\tanh$ is employed as the activation function. The Adam~\cite{kingma2014adam} optimizer is utilized with an initial learning rate of $0.01$, gradually descending according to the CosineAnnealingLR method. A warm start is applied with $500$ epochs, and the weight decay is set to $1\times 10^{-4}$. $K_{ade}=K_{data}=5,000$ points are randomly selected from each MRI image data obtained at different time points for training the network. The weight for the ADE is set to $w_{ade}=100$, while the weight for the measured data is set to $w_{data}=1$. The network is trained for a total of $30,000$ epochs.

\begin{figure}[!t]
\centerline{\includegraphics[width=0.7\columnwidth]{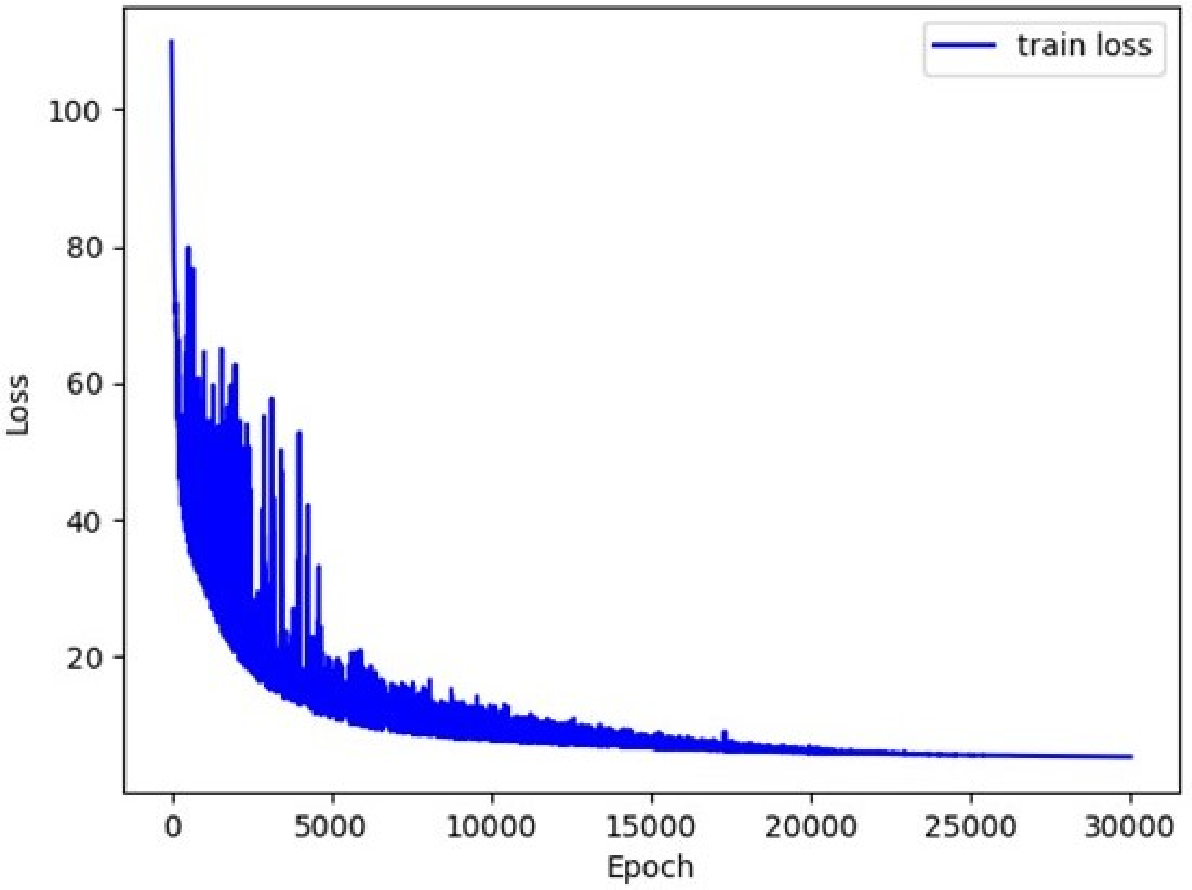}}
\caption{The loss values change with epochs during the training of the network on the dataset-1.}\label{loss-N5-20}
\end{figure}

\begin{figure*}[!t]
\centering
\subfigure[]{
\label{D-N5-20}
\includegraphics[width=0.7\columnwidth]{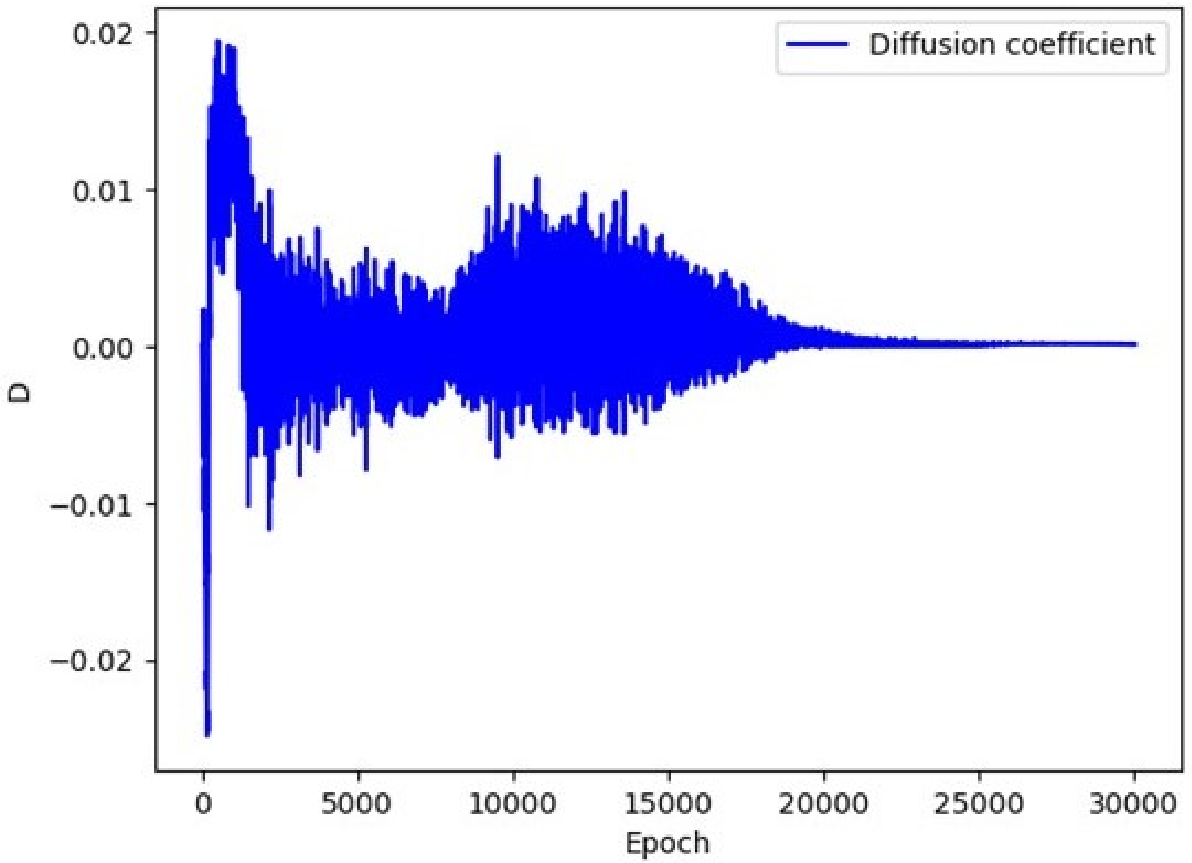}}
\quad
\subfigure[]{
\label{V-N5-20}
\includegraphics[width=0.7\columnwidth]{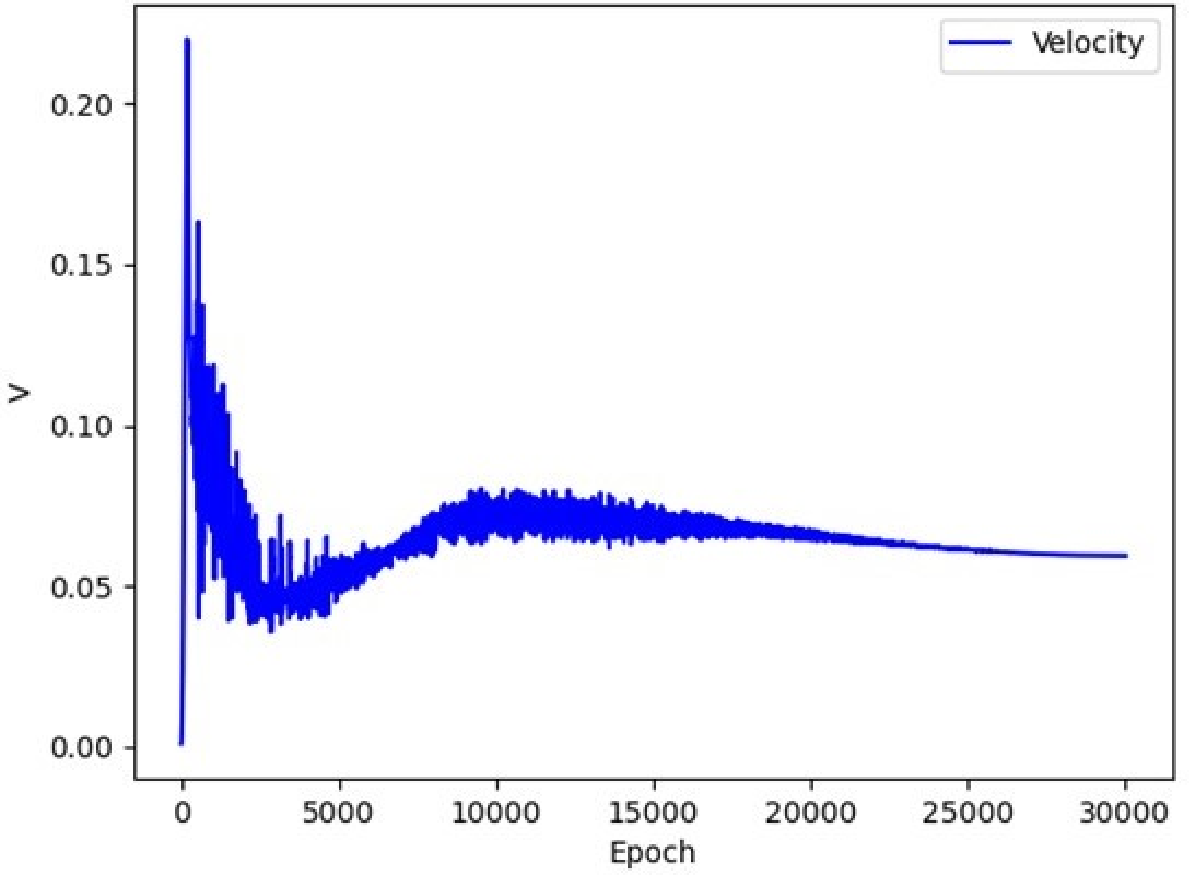}}
\caption{The evolving values of the diffusion coefficient $D$ and velocity $v$ during the training of the network on the dataset-1. (a) The evolution of the diffusion coefficient $D$ values during the training of the network on the dataset-1. (b) The evolution of the velocity $v$ values during the training of the network on the dataset-1.}
\label{DV-N5-20}
\end{figure*}

\begin{figure*}[!t]
\centering
\subfigure[]{
\label{pred-20min-10-N5-20}
\includegraphics[width=4cm]{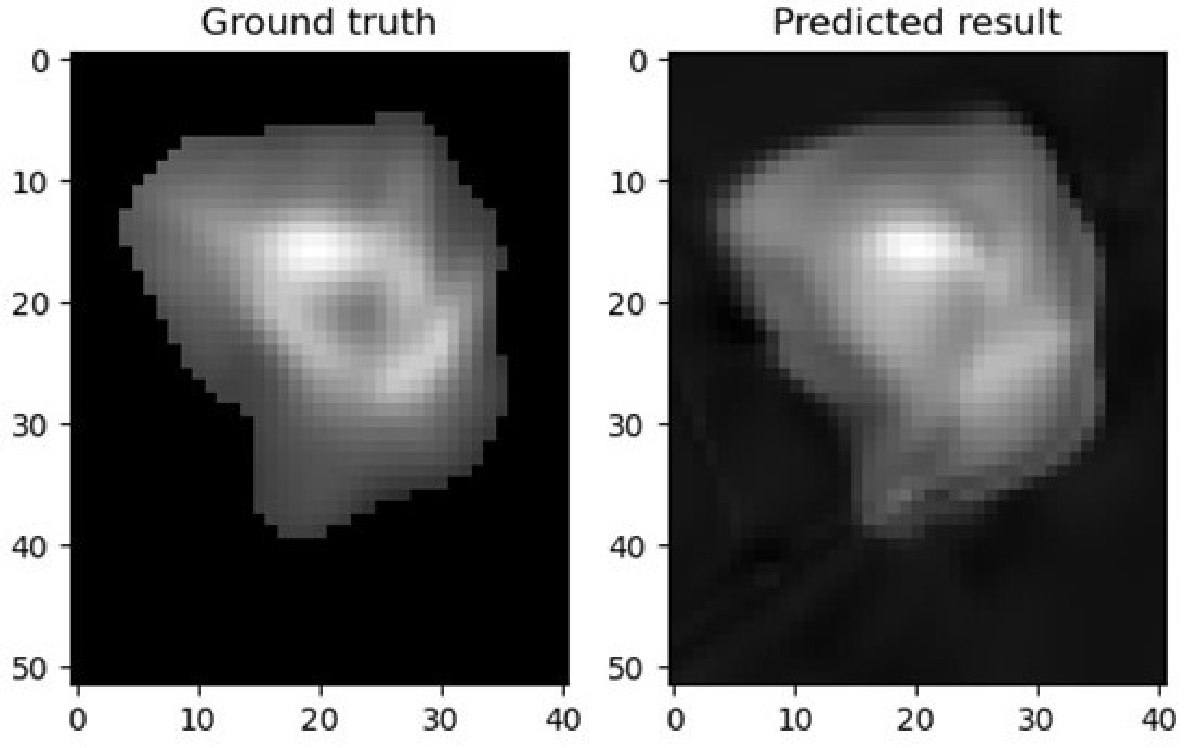}}
\quad
\subfigure[]{
\label{pred-20min-15-N5-15}
\includegraphics[width=4cm]{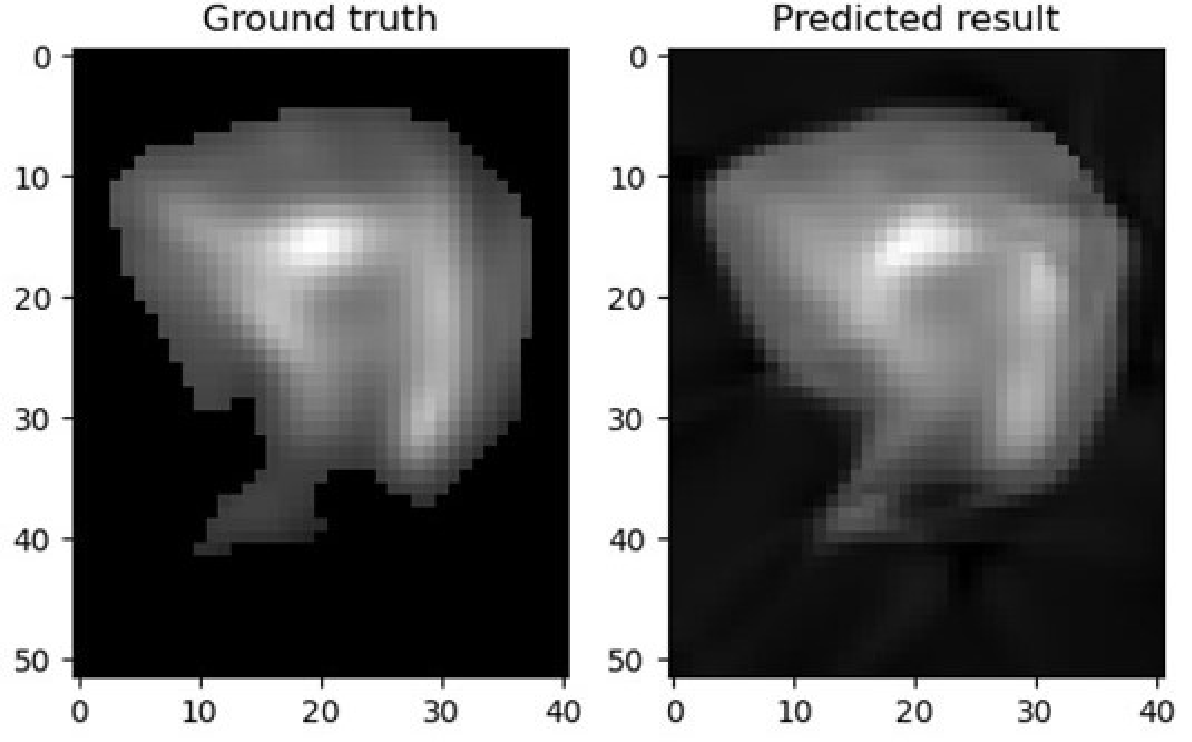}}
\quad
\subfigure[]{
\label{pred-20min-20-N5-20}
\includegraphics[width=4cm]{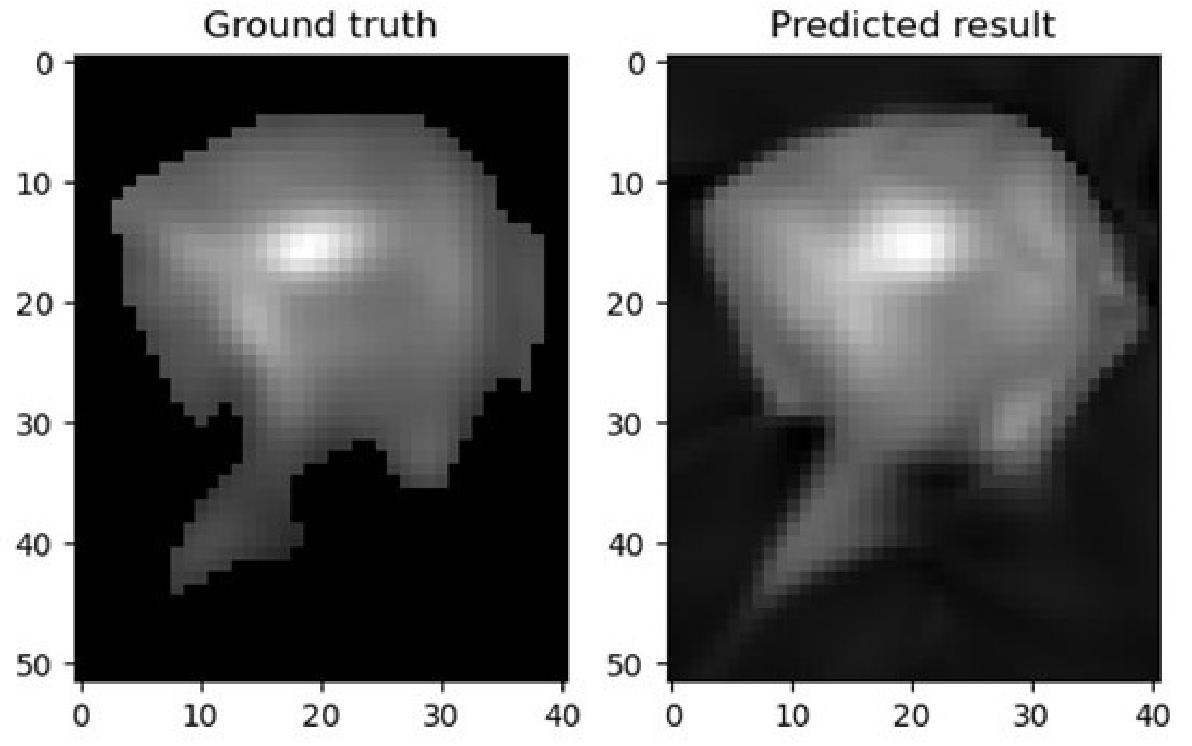}}
\quad
\subfigure[]{
\label{pred-20min-25-N5-20}
\includegraphics[width=4cm]{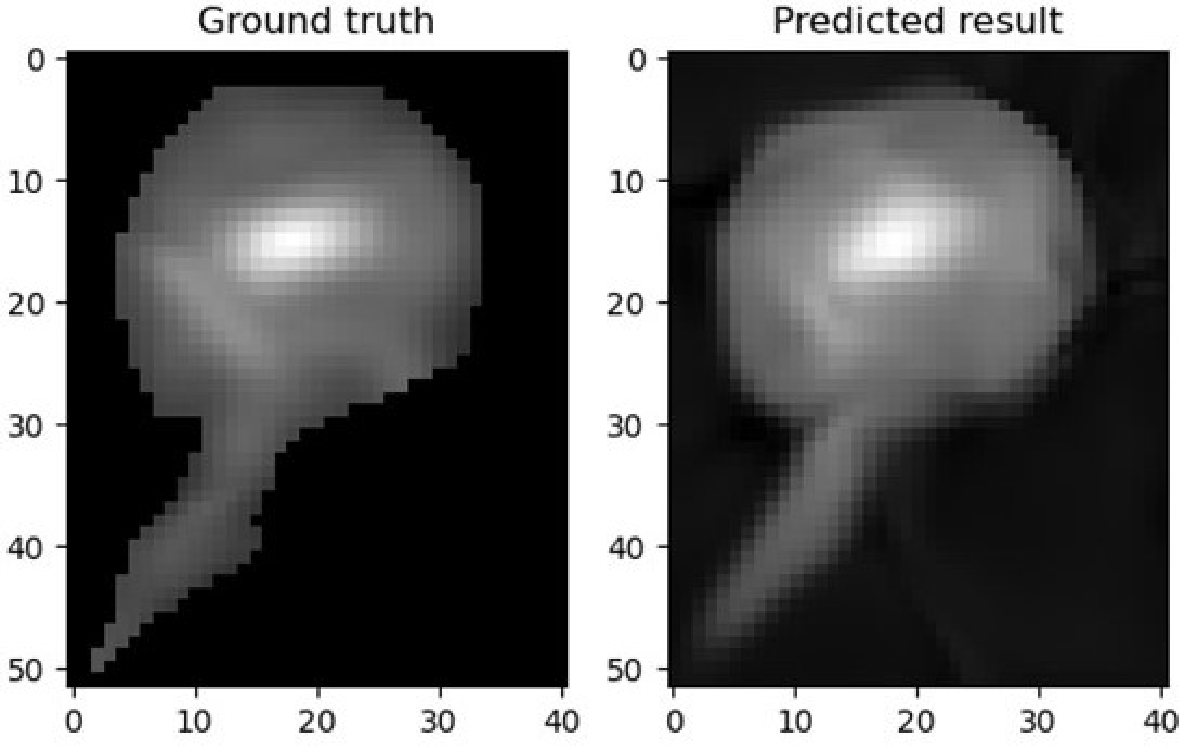}}
\caption{Comparison of the ground truth (coronal plane) and predicted results for the dataset-1 at $t=20$ min. The corresponding mean squared error between the ground truth and the prediction is $3.79\times10^{-2}$. (a)-(d) denote the Ground truth and prediction for the slices $10$, $15$, $20$ and $25$, respectively.}
\label{pred-20min-dataset1}
\end{figure*}

\begin{figure*}[!t]
\centering
\subfigure[]{
\label{pred-50min-10-N5-20}
\includegraphics[width=4cm]{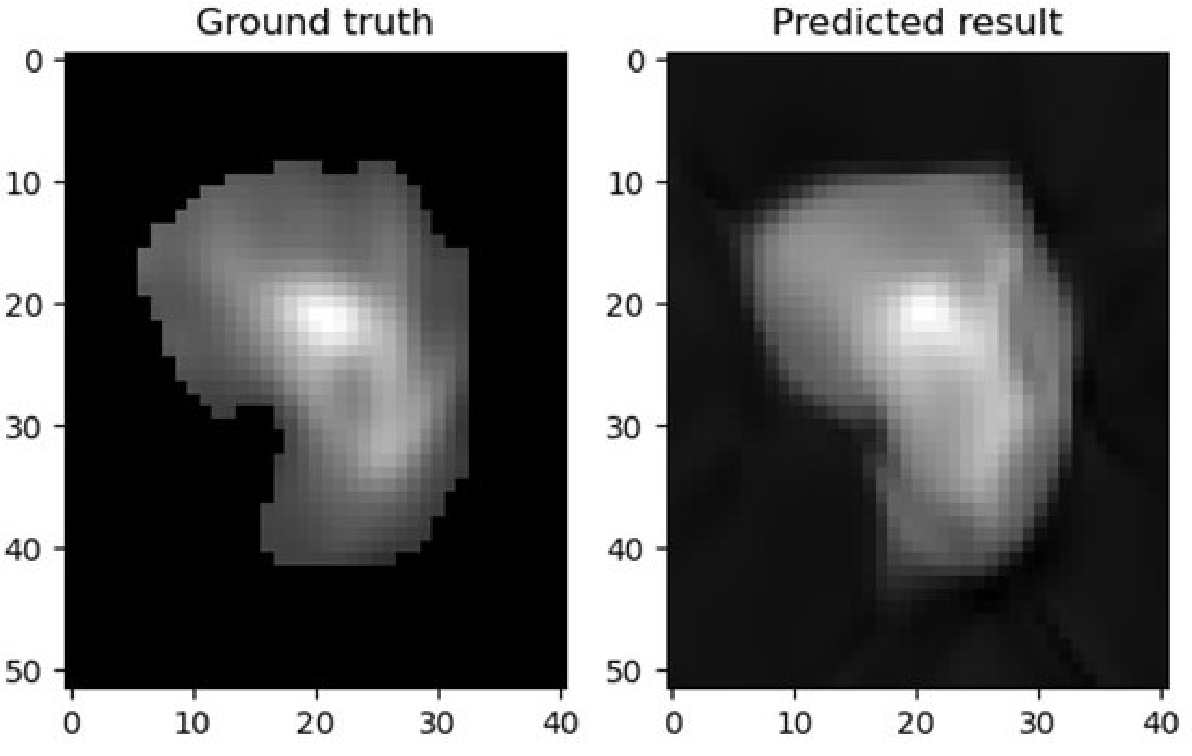}}
\quad
\subfigure[]{
\label{pred-50min-15-N5-20}
\includegraphics[width=4cm]{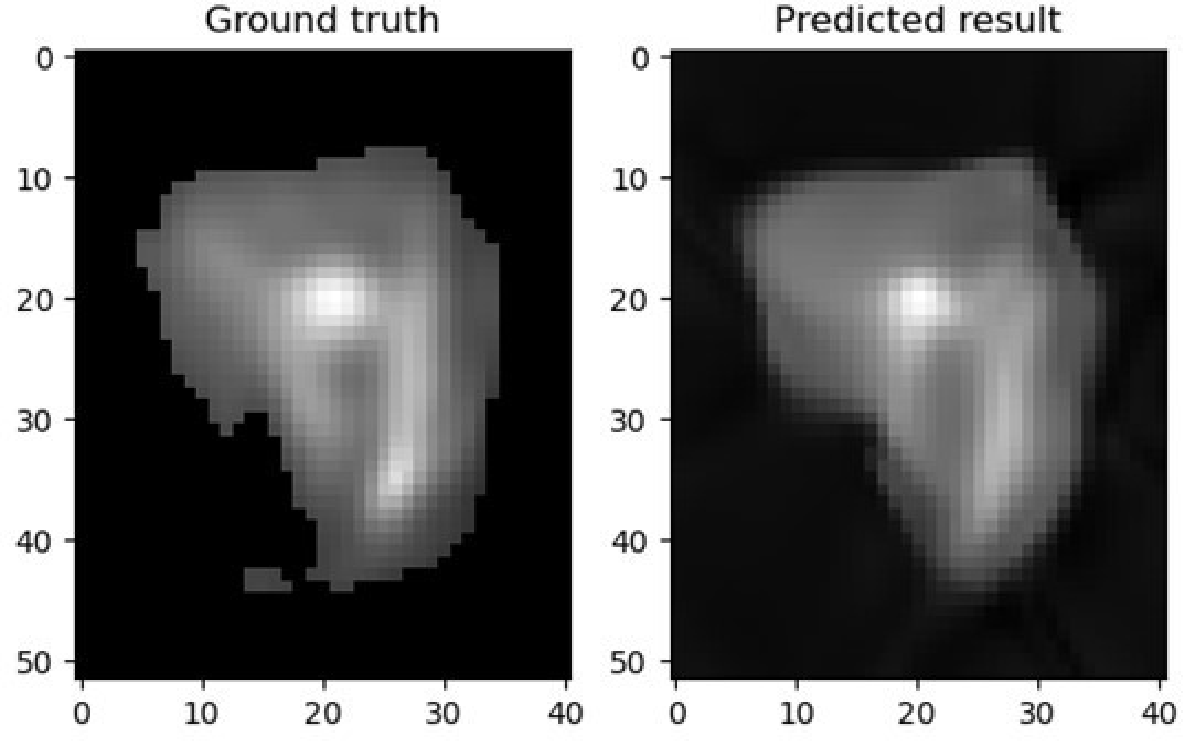}}
\quad
\subfigure[]{
\label{pred-50min-20-N5-20}
\includegraphics[width=4cm]{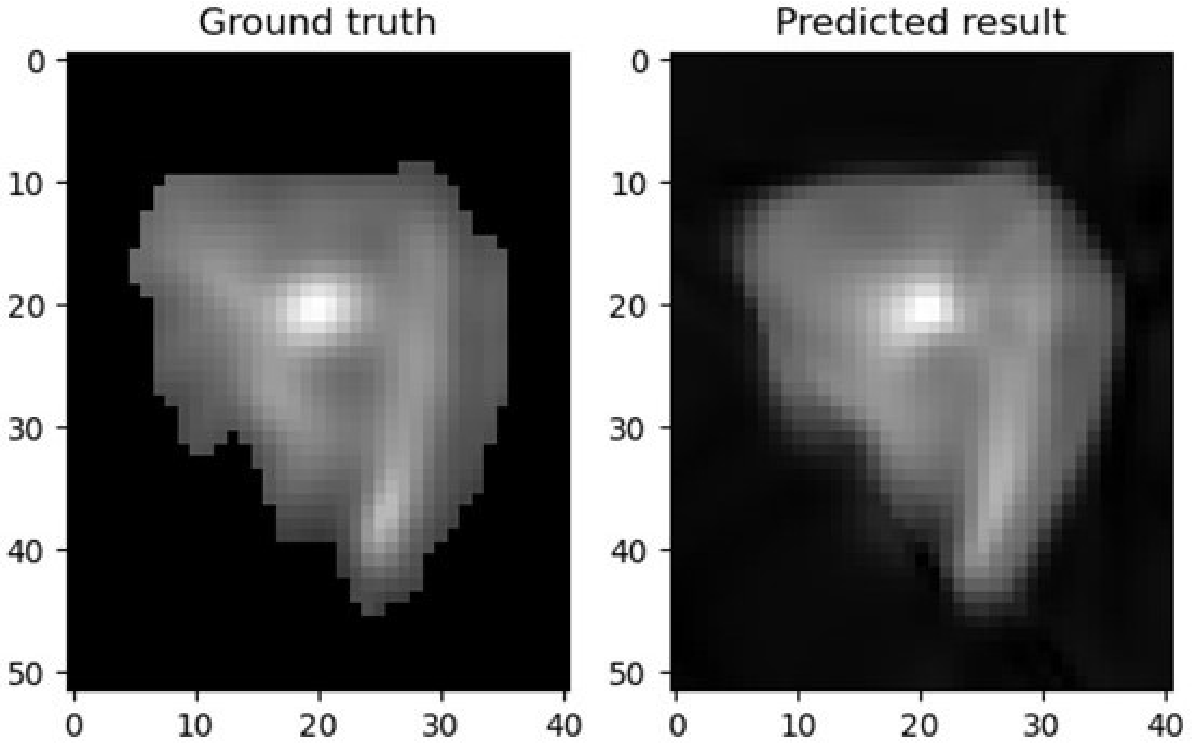}}
\quad
\subfigure[]{
\label{pred-50min-25-N5-20}
\includegraphics[width=4cm]{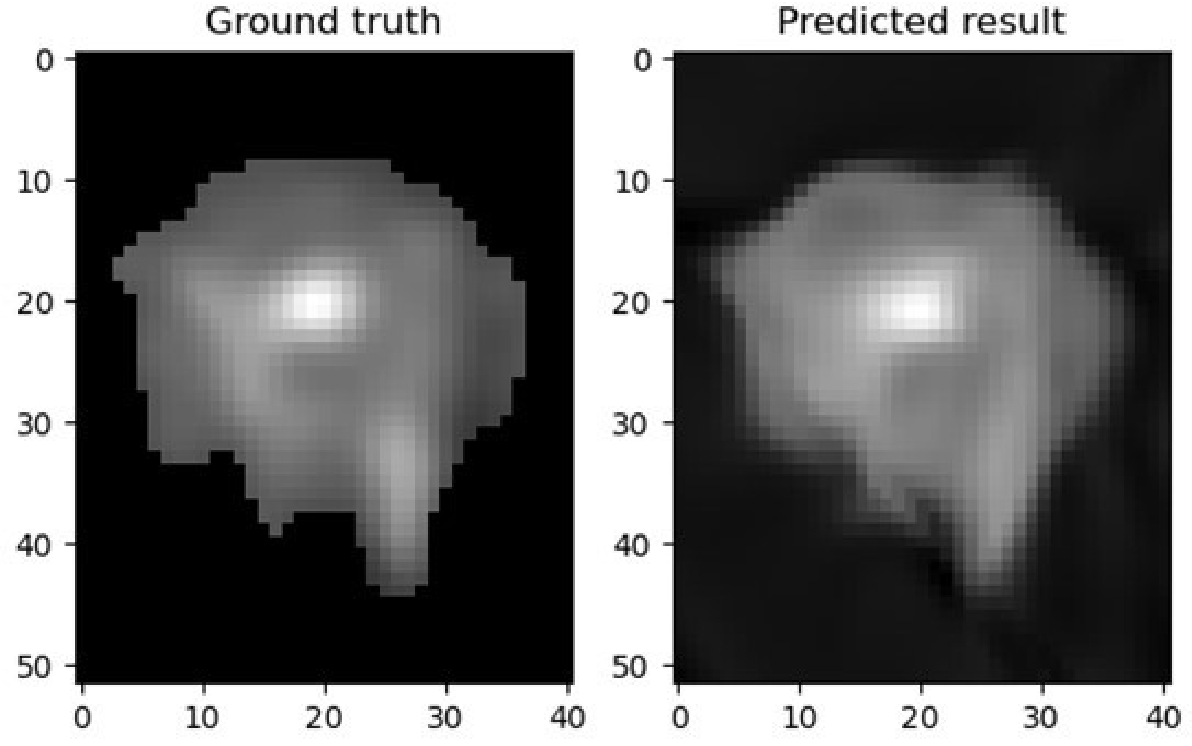}}
\caption{Comparison of the ground truth (coronal plane) and predicted results for the dataset-1 at $t=50$ min. The corresponding mean squared error between the ground truth and the prediction is $4.49\times10^{-2}$. (a)-(d) denote the Ground truth and prediction for the slices $10$, $15$, $20$ and $25$, respectively.}
\label{pred-50min-dataset1}
\end{figure*}

\begin{figure*}[!t]
\centering
\subfigure[]{
\label{pred-70min-10-N5-20}
\includegraphics[width=4cm]{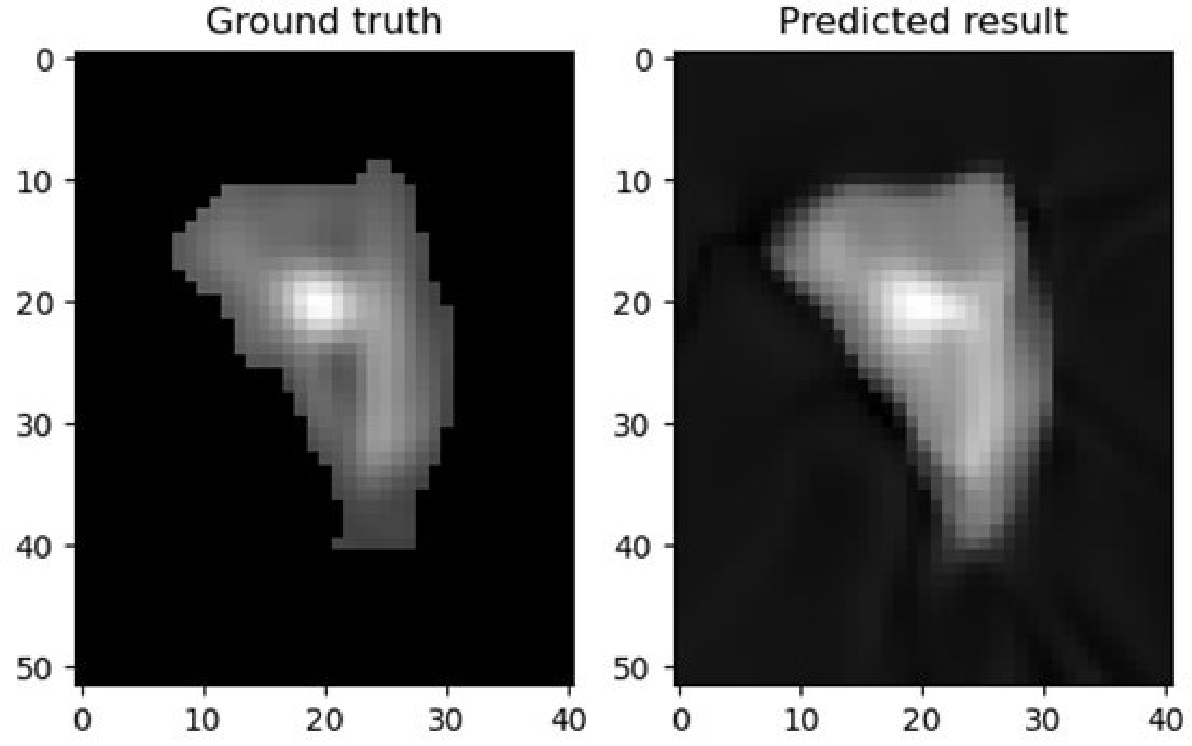}}
\quad
\subfigure[]{
\label{pred-70min-15-N5-20}
\includegraphics[width=4cm]{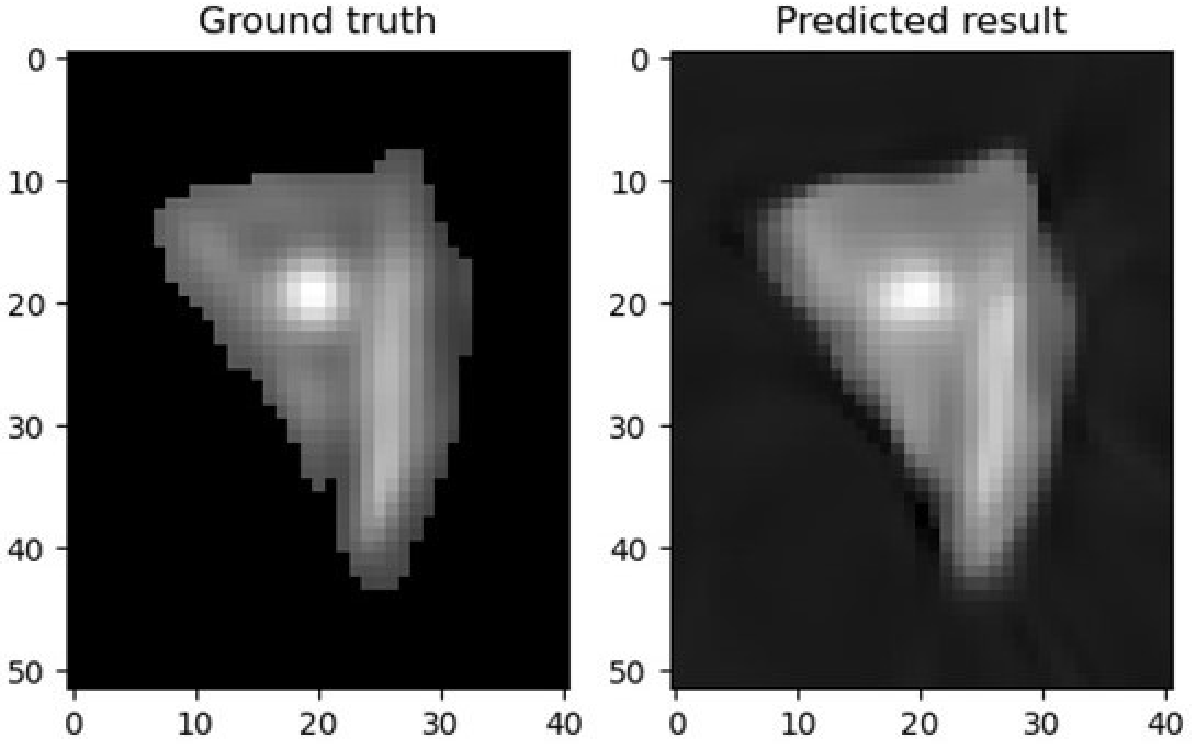}}
\quad
\subfigure[]{
\label{pred-70min-20-N5-20}
\includegraphics[width=4cm]{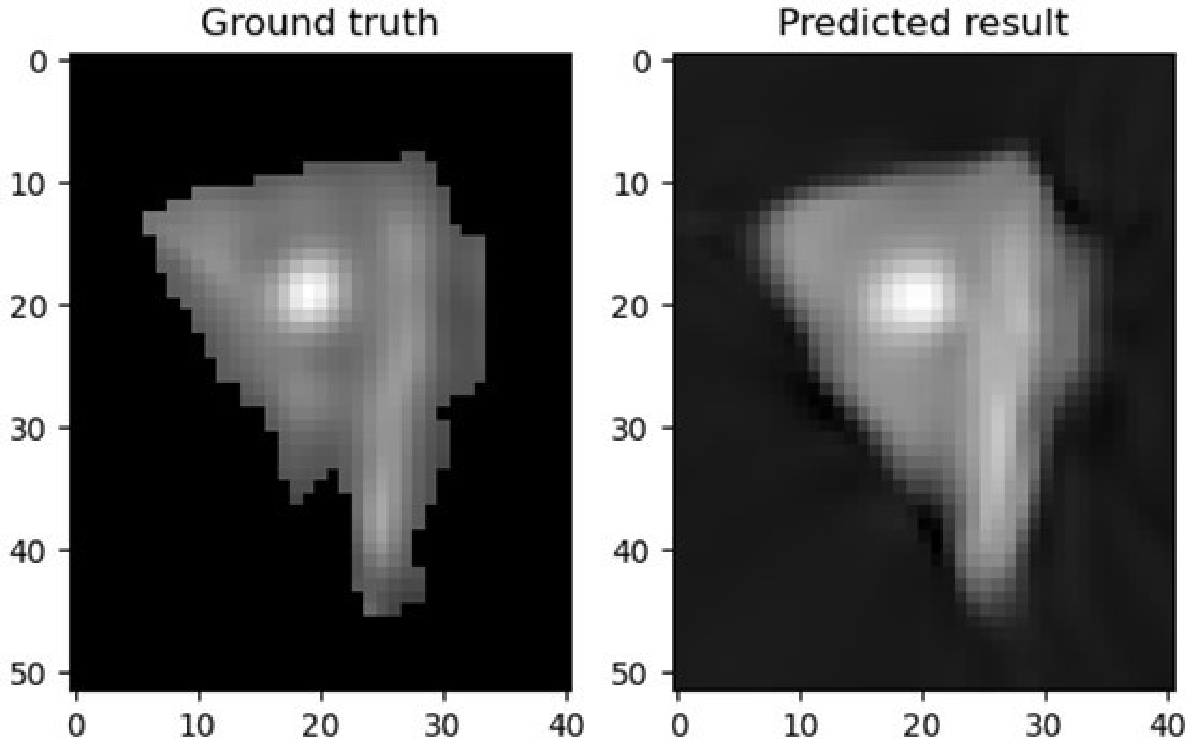}}
\quad
\subfigure[]{
\label{pred-70min-25-N5-20}
\includegraphics[width=4cm]{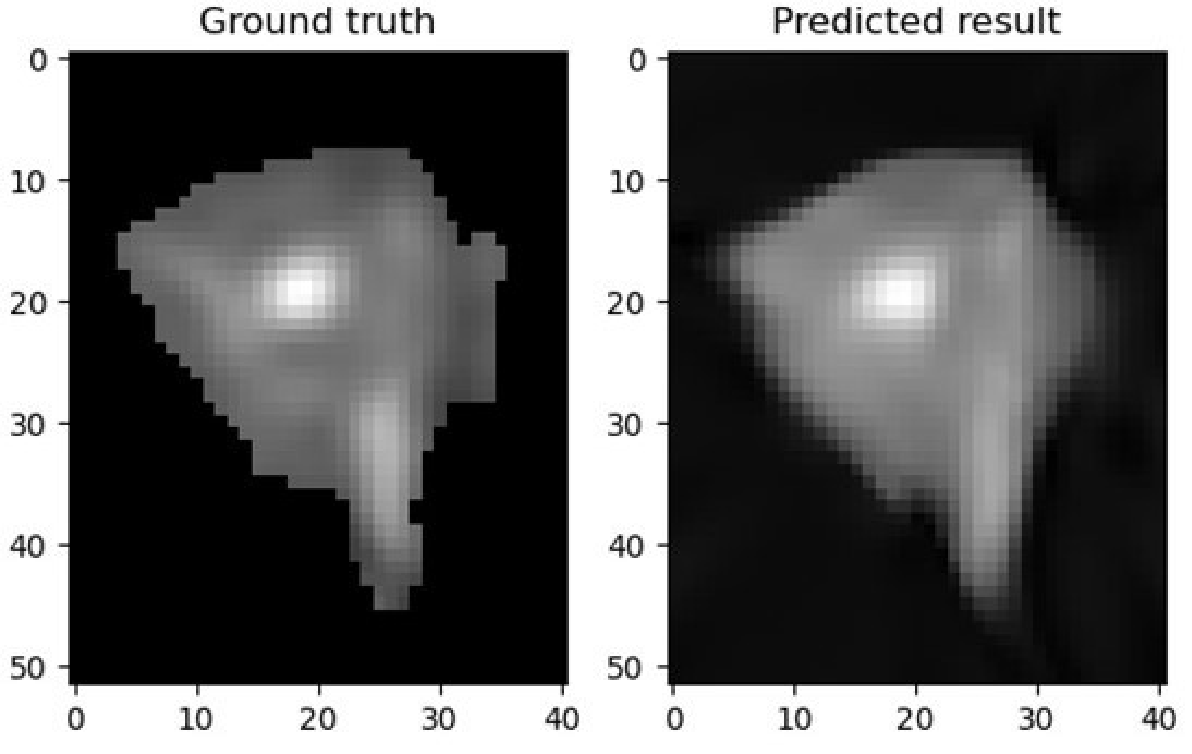}}
\caption{Comparison of the ground truth (coronal plane) and predicted results for the dataset-1 at $t=70$ min. The corresponding mean squared error between the ground truth and the prediction is $5.57\times10^{-2}$. (a)-(d) denote the Ground truth and prediction for the slices $10$, $15$, $20$ and $25$, respectively.}
\label{pred-70min-dataset1}
\end{figure*}

The evolution of loss values during network training on the dataset-1 is illustrated in Fig.~\ref{loss-N5-20}, indicating a swift convergence. Fig.~\ref{D-N5-20} and~\ref{V-N5-20} illustrate the changes in the values of the diffusion coefficient $D$ and the velocity $v$ throughout network training, respectively. Initialized with $D_0=1.00\times10^{-4}$ mm$^2$/s and $v_0=1.00\times10^{-3}$ mm/s, the values of $D$ and $v$ initially oscillate and eventually converge to stable values. The final estimated diffusion coefficient $D$ is $1.25\times10^{-4}$ mm$^2$/s, and the velocity $v$ is $5.95\times10^{-2}$ mm/s. The predicted results of PINN and the ground truth which are acquired data at $20$ mins, $50$ mins and $70$ mins are presented simultaneously in Fig.~\ref{pred-20min-dataset1},~\ref{pred-50min-dataset1} , and~\ref{pred-70min-dataset1}, respectively. Additionally, the corresponding mean squared errors (MSEs) between the ground truth and the predictions are $3.79\times10^{-2}$, $4.49\times10^{-2}$ and $5.57\times10^{-2}$, respectively. Consequently, these experimental results demonstrate that PINN provides accurate solutions to problem~(\ref{eq1-2}) compared to the ground truth.

%

\subsection{Results on the Dataset-2}

On dataset-2, the network parameters (i.e., weights and biases) are initialized using the Glorot initialization method, and $\tanh$ is employed as the activation function. Adam~\cite{kingma2014adam} serves as the optimizer, with an initial learning rate set at $0.01$ that gradually descends according to the CosineAnnealingLR method. A warm start is applied, spanning $500$ epochs. The weight decay is set to $1 \times 10^{-4}$. $K_{ade}=K_{data}=5,000$ points are randomly selected from each MRI image data obtained at different time points for training the network. The weight for the ADE is set to $w_{ade}=500$, while the weight for the measured data is set to $w_{data}=1$. The network undergoes training for a total of $30,000$ epochs.

\begin{figure}[!t]
\centerline{\includegraphics[width=0.7\columnwidth]{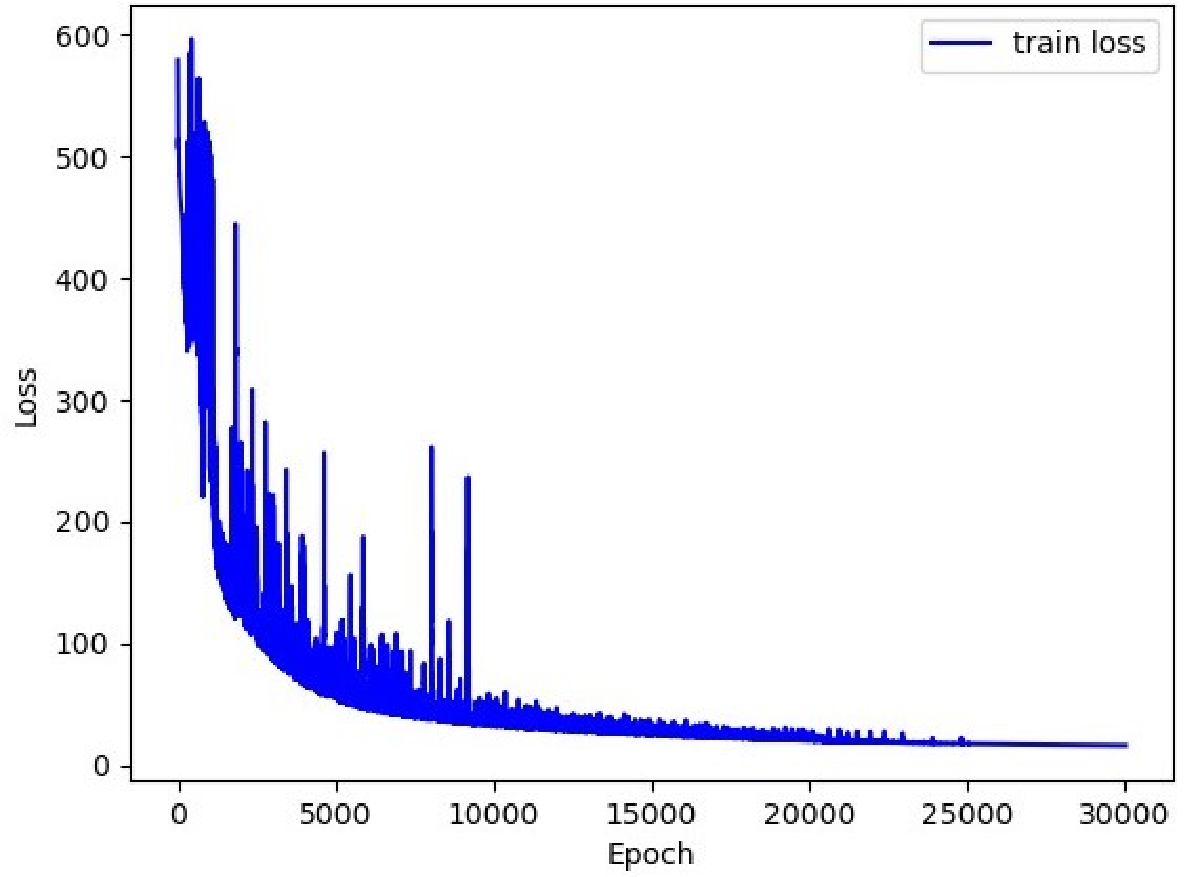}}
\caption{The loss values change with epochs during the training of the network on the dataset-2.}\label{loss-N8-30}
\end{figure}

\begin{figure*}[!t]
\centering
\subfigure[]{
\label{D-old}
\includegraphics[width=0.7\columnwidth]{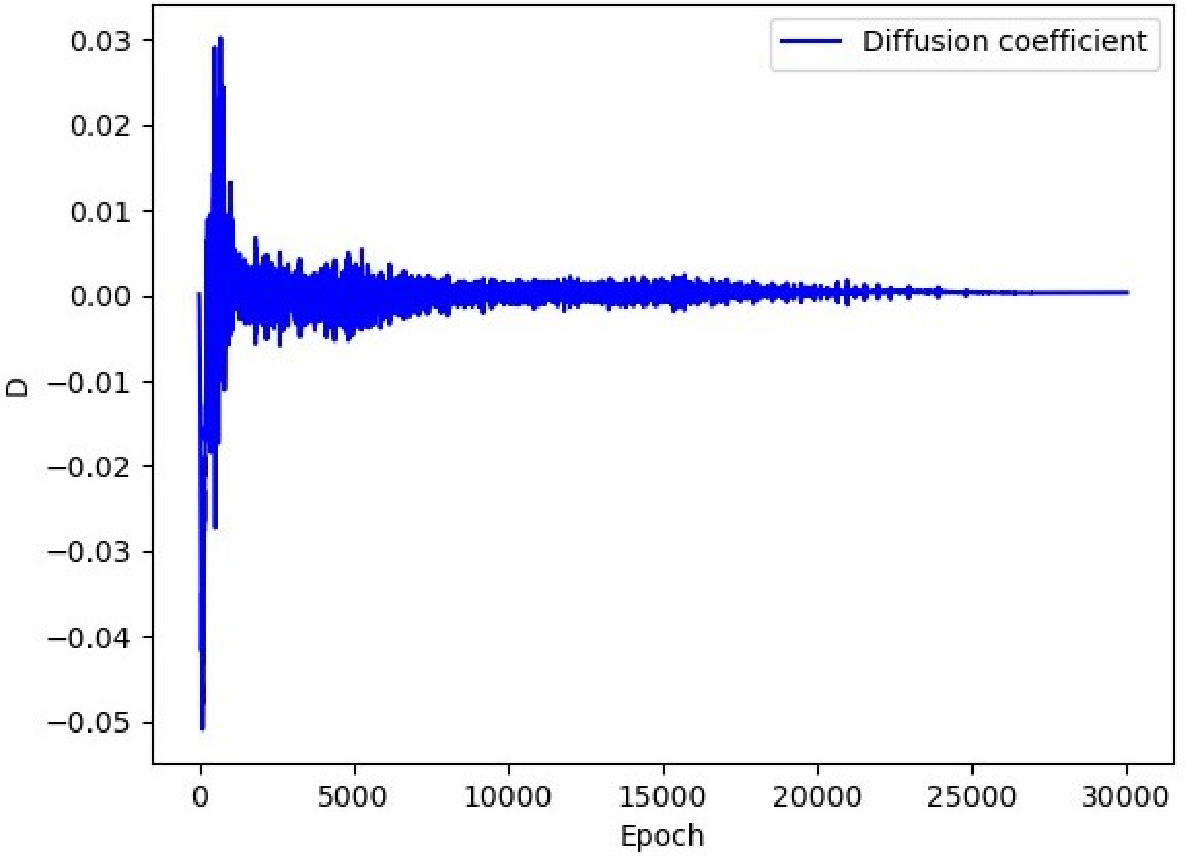}}
\quad
\subfigure[]{
\label{V-old}
\includegraphics[width=0.7\columnwidth]{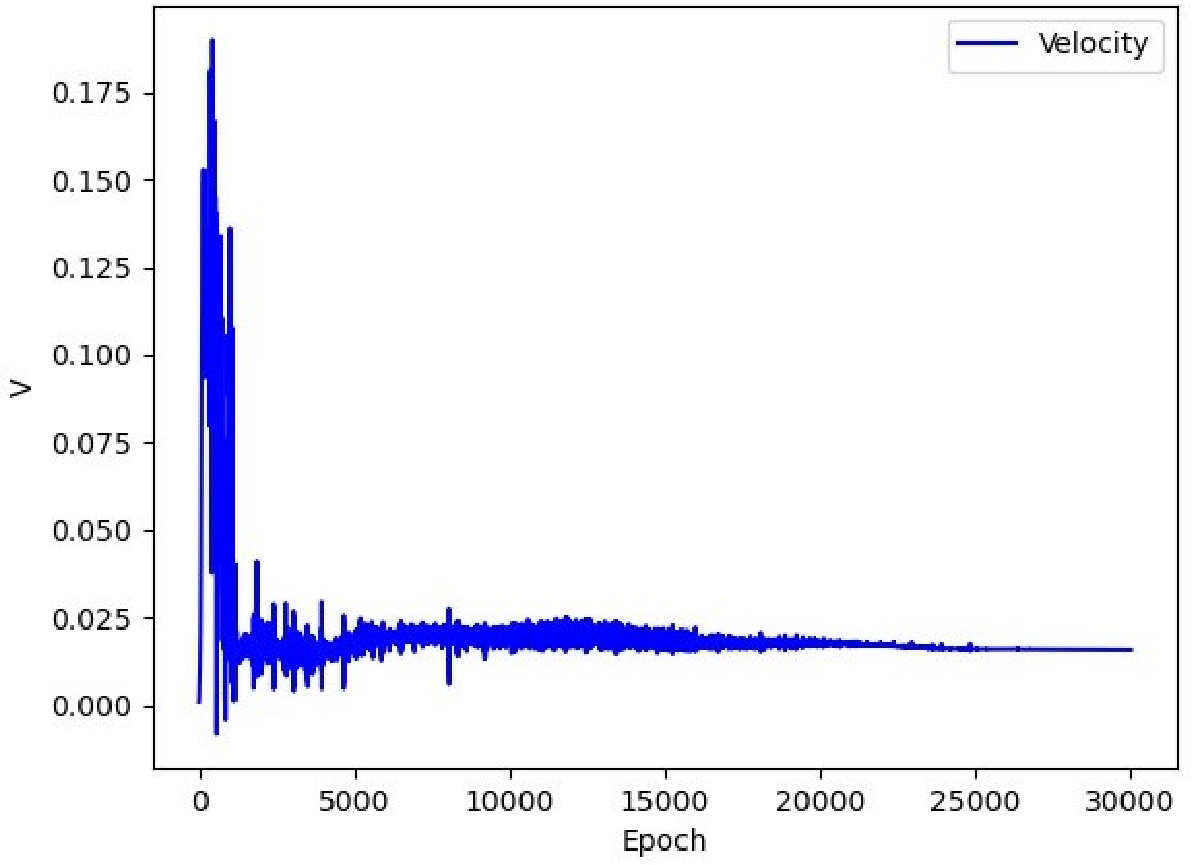}}
\caption{The evolving values of the diffusion coefficient $D$ and velocity $v$ during the training of the network on the dataset-2. (a) The evolution of the diffusion coefficient $D$ values during the training of the network on the dataset-2. (b) The evolution of the velocity $v$ values during the training of the network on the dataset-2.}
\label{DV-old}
\end{figure*}

\begin{figure*}[!t]
\centering
\subfigure[]{
\label{pred-30min-5-N8-30}
\includegraphics[width=4cm]{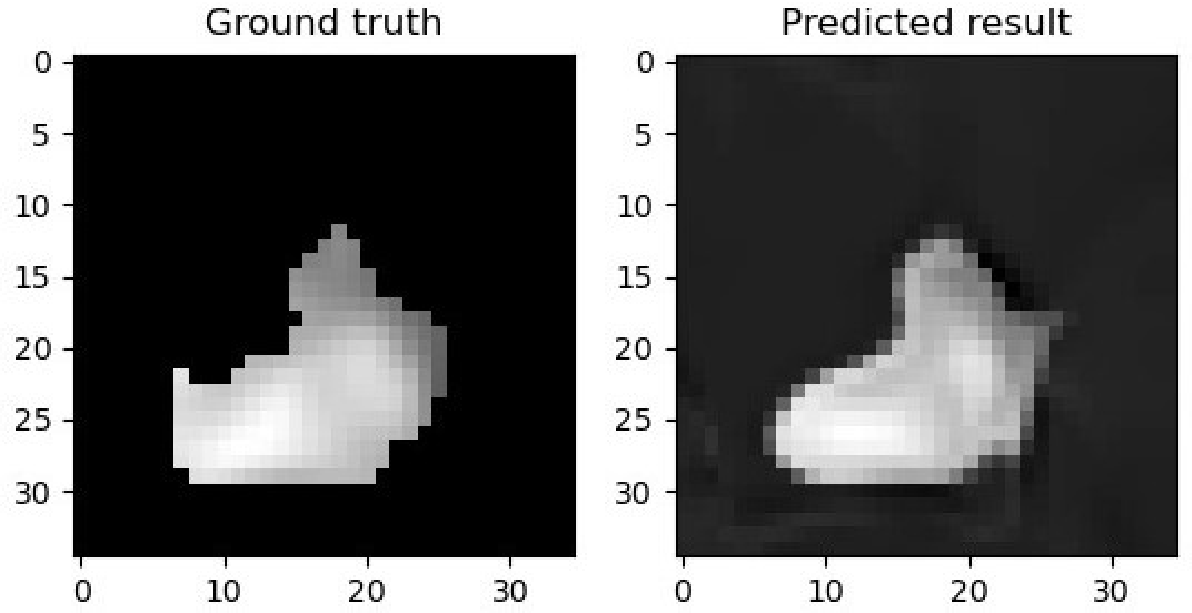}}
\quad
\subfigure[]{
\label{pred-30min-8-N8-30}
\includegraphics[width=4cm]{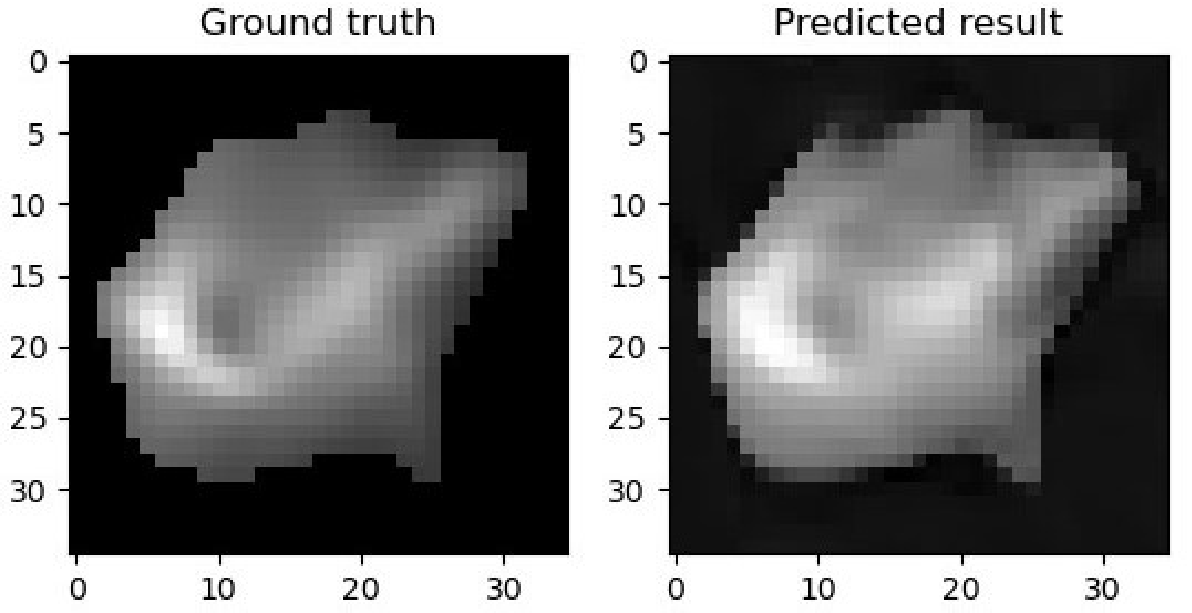}}
\quad
\subfigure[]{
\label{pred-30min-10-N8-30}
\includegraphics[width=4cm]{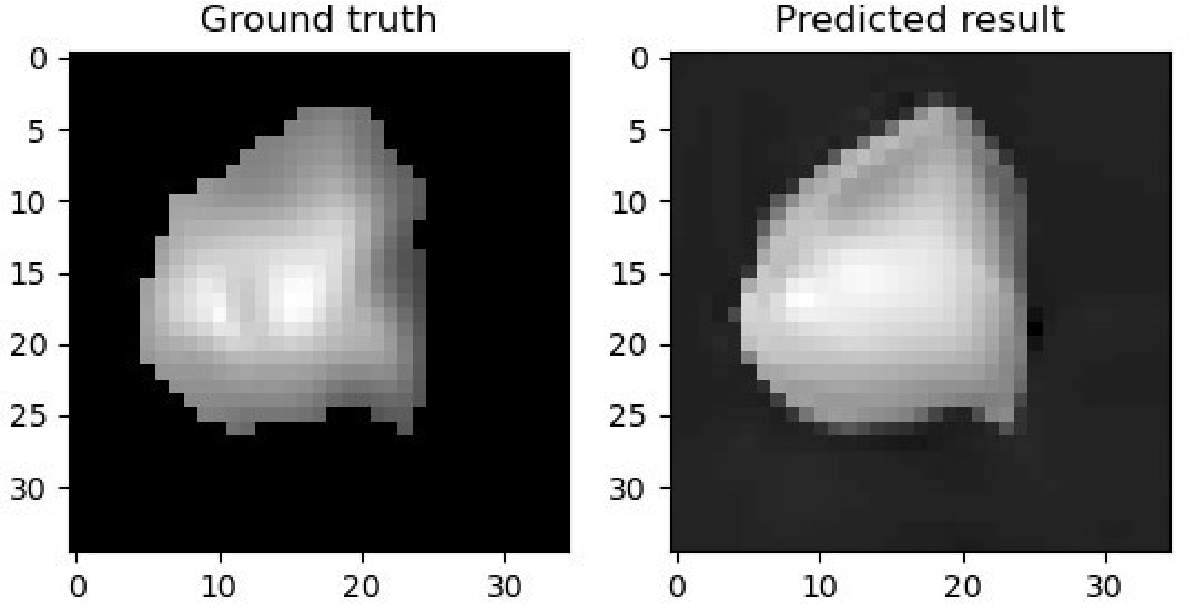}}
\quad
\subfigure[]{
\label{pred-30min-12-N8-30}
\includegraphics[width=4cm]{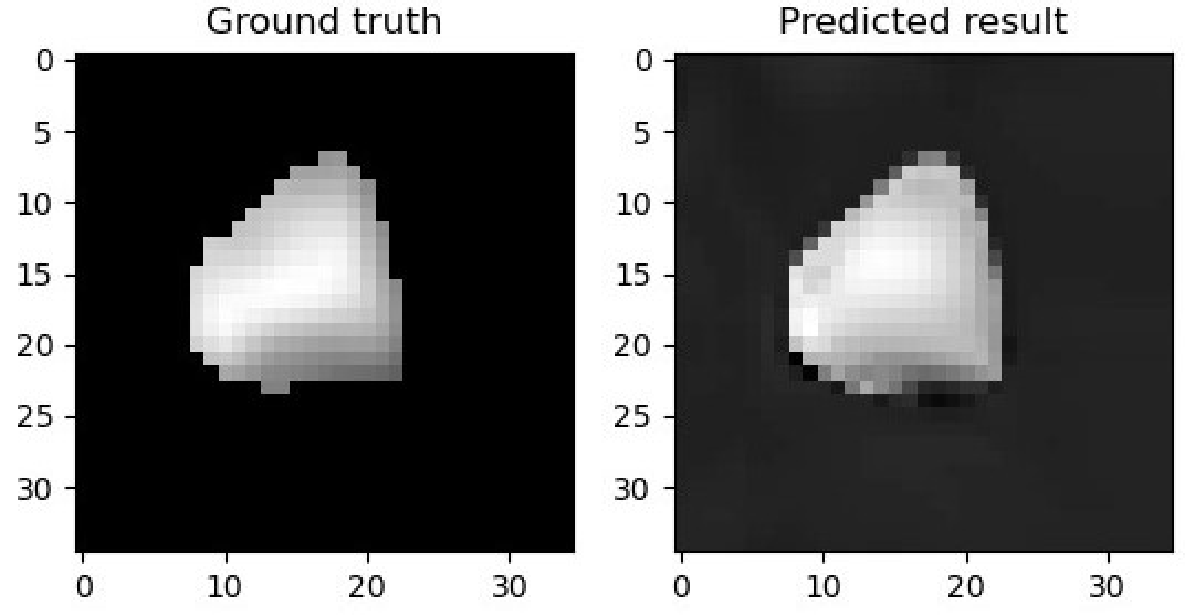}}
\caption{Comparison of the ground truth (coronal plane) and predicted results for the dataset-2 at $t=30$ min. The corresponding mean squared error between the ground truth and the prediction is $2.53\times10^{-2}$. (a)-(d) denote the ground truth and prediction for the slice $5$, $8$, $10$ and $12$, respectively.}
\label{pred-30min-N8-30}
\end{figure*}

\begin{figure*}[!t]
\centering
\subfigure[]{
\label{pred-50min-5-N8-30}
\includegraphics[width=4cm]{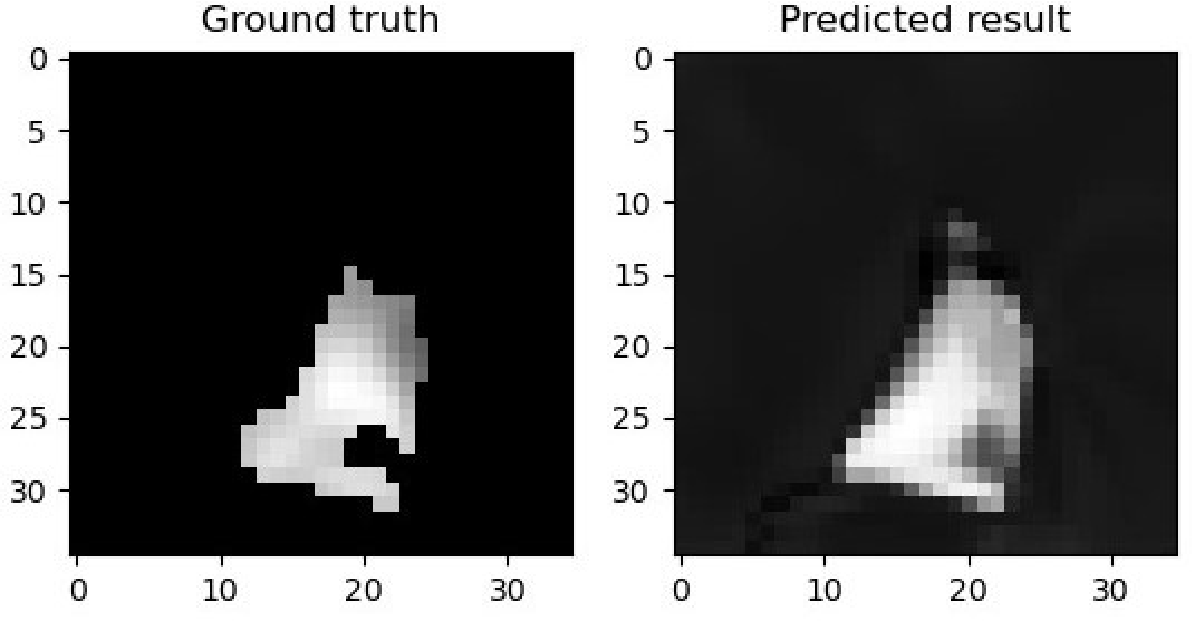}}
\quad
\subfigure[]{
\label{pred-50min-5-N8-30}
\includegraphics[width=4cm]{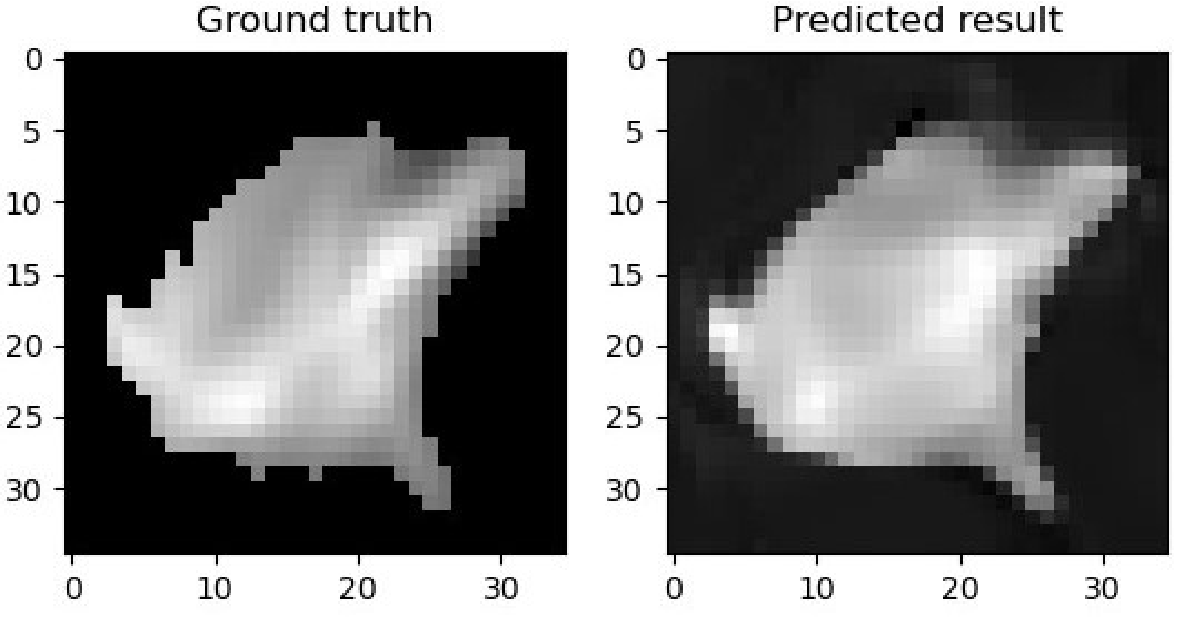}}
\quad
\subfigure[]{
\label{pred-50min-10-N8-30}
\includegraphics[width=4cm]{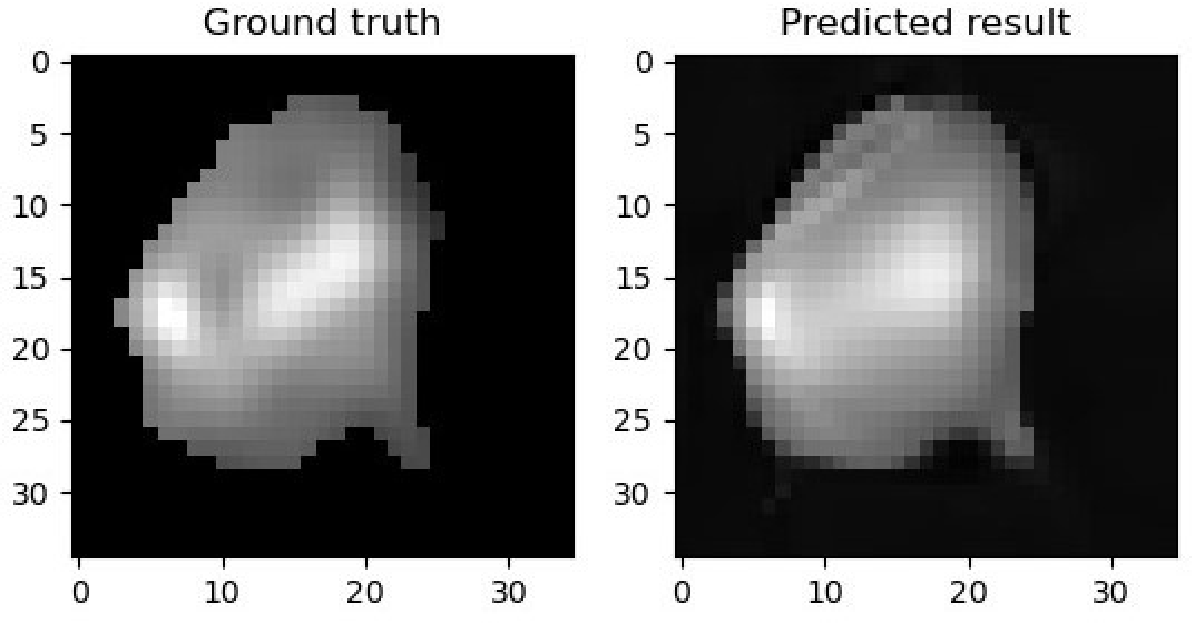}}
\quad
\subfigure[]{
\label{pred-50min-12-N8-30}
\includegraphics[width=4cm]{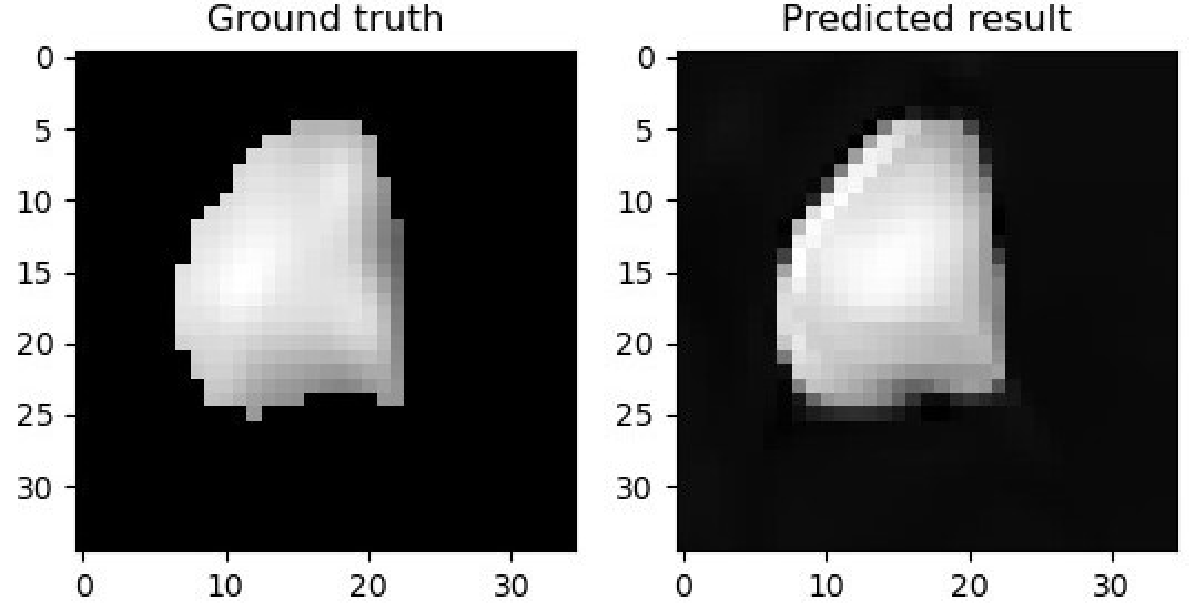}}
\caption{Comparison of the ground truth (coronal plane) and predicted results for the dataset-2 at $t=50$ min. The corresponding mean squared error between the ground truth and the prediction is $3.17\times10^{-2}$. (a)-(d) denote the ground truth and prediction for the slice $5$, $8$, $10$ and $12$, respectively.}
\label{pred-50min-N8-30}
\end{figure*}

\begin{figure*}[!t]
\centering
\subfigure[]{
\label{pred-90min-5-N8-30}
\includegraphics[width=4cm]{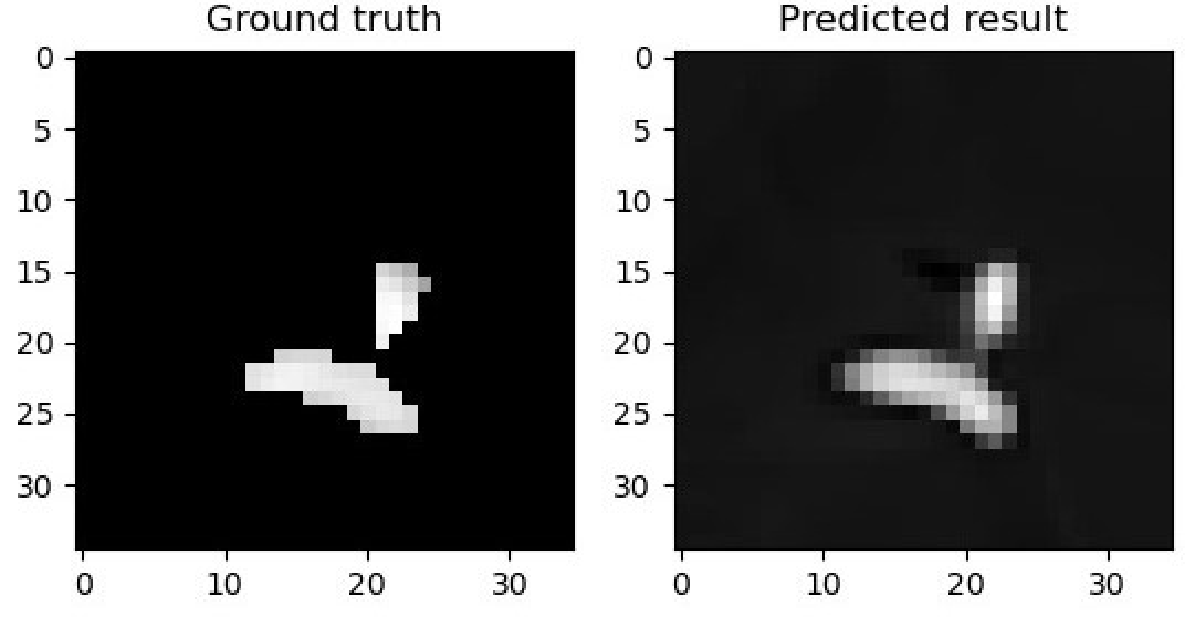}}
\quad
\subfigure[]{
\label{pred-90min-5-N8-30}
\includegraphics[width=4cm]{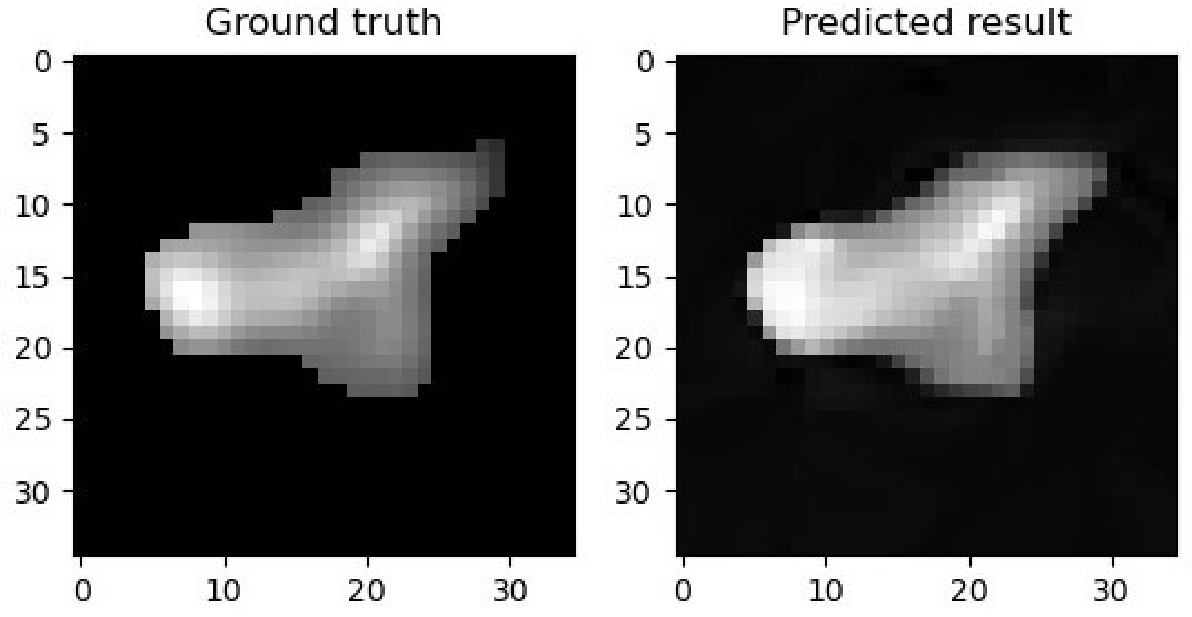}}
\quad
\subfigure[]{
\label{pred-90min-10-N8-30}
\includegraphics[width=4cm]{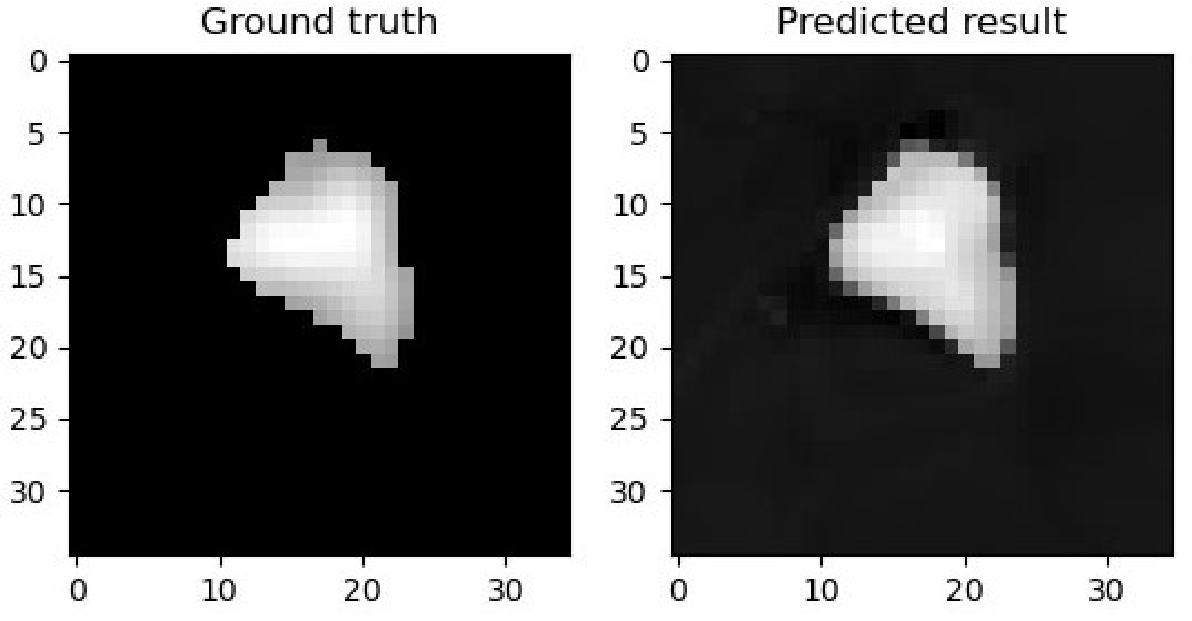}}
\quad
\subfigure[]{
\label{pred-90min-12-N8-30}
\includegraphics[width=4cm]{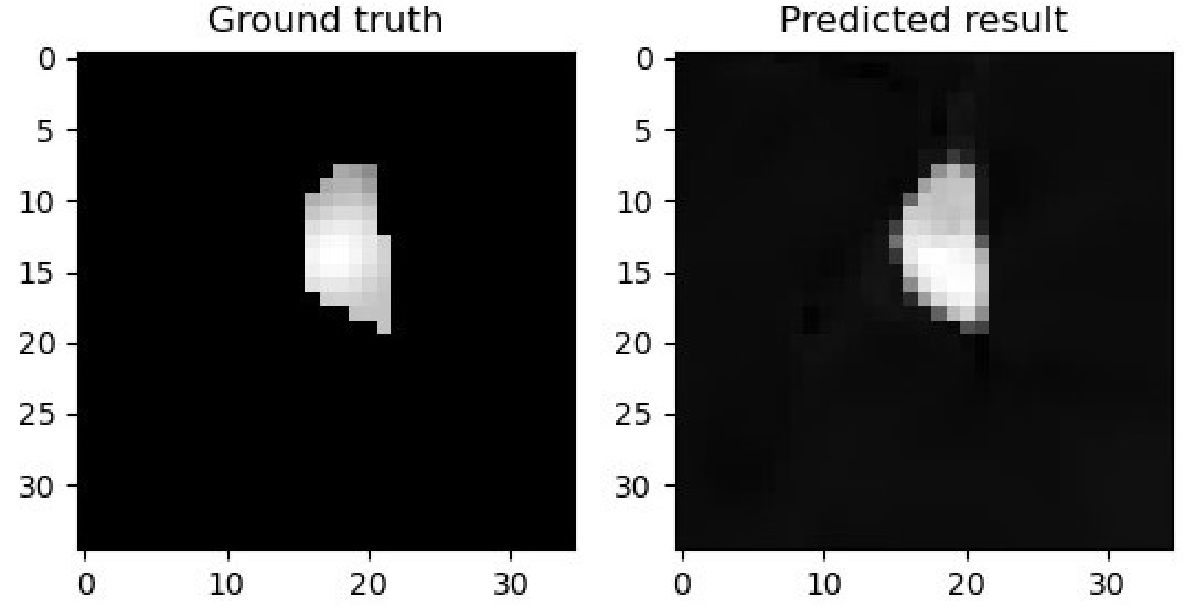}}
\caption{Comparison of the ground truth (coronal plane) and predicted results for the dataset-2 at $t=90$ min. The corresponding mean squared error between the ground truth and the prediction is $2.95\times10^{-2}$. (a)-(d) denote the ground truth and prediction for the slice $5$, $8$, $10$ and $12$, respectively.}
\label{pred-90min-N8-30}
\end{figure*}

The evolution of loss values throughout the network training on dataset-2 is depicted in Fig.~\ref{loss-N8-30}, indicating a rapid convergence. Figs.~\ref{D-old} and~\ref{V-old} display the changes in the values of the diffusion coefficient $D$ and velocity $v$ during network training, respectively. Both parameters exhibit initial oscillations followed by convergence to stable values. The final estimated diffusion coefficient $D$ is $3.11 \times 10^{-4}$ mm$^2$/s, and the velocity $v$ is $1.57 \times 10^{-2}$ mm/s. Figs.~\ref{pred-30min-N8-30},~\ref{pred-50min-N8-30}, and~\ref{pred-90min-N8-30} illustrate the ground truth-prediction comparisons at times of $30$ mins, $50$ mins, and $90$ mins, respectively. Additionally, the corresponding MSEs between the ground truth and predictions are $2.53 \times 10^{-2}$, $3.17 \times 10^{-2}$, and $2.95 \times 10^{-2}$, respectively. These results collectively affirm that PINN consistently provides accurate approximations for solving the problem~(\ref{eq1-2}) and predicting molecular concentration with high precision.

%

\subsection{Comparison with the Existing Method}

To further assess the effectiveness of the proposed method, a comparison is conducted with the existing method proposed by Han et al. in~\cite{han2014novel}, and the results are presented in Table~\ref{comp_dv}.

The method presented in~\cite{han2014novel} explored molecular transport within the ECS by exclusively focusing on the diffusion term, while neglecting the advection term in the advection-diffusion equation. This approach is less precise in describing the molecular transport within the ECS. Nevertheless, the calculated results of $D$ from \cite{han2014novel} can still serve as a reference. The comparison results in Table~\ref{comp_dv} indicate that the calculated values of $D$ by both the method in \cite{han2014novel} and the proposed method are approximately in the same order of magnitude, affirming the effectiveness of the proposed approach. Notably, the proposed method can automatically obtain the velocity $v$, a parameter that cannot be calculated by the method in \cite{han2014novel}. Recognizing the importance of considering both diffusion and advection terms for characterizing the molecular transport within the ECS, the proposed method demonstrates a clear advantage over the existing method in \cite{han2014novel}.


\begin{table}[!t]
\renewcommand{\arraystretch}{1.5}
\setlength{\tabcolsep}{3pt}
\caption{Comparison of $D$ and $v$ on the two datasets with the existing method}
\label{comp_dv}
\begin{center}
\begin{tabular}{|c|c|c|c|c|}
\hline
\multirow{3}{*}{\bf Methods} & \multicolumn{2}{c|}{\bf Dataset-1} & \multicolumn{2}{c|}{\bf Dataset-2}\\
\cline{2-5}
& $D$ (mm$^2$/s) & $v$ (mm/s) & $D$ (mm$^2$/s) & $v$ (mm/s) \\
\hline
\cite{han2014novel}      & $4.94\times10^{-4}$               & -                                  & $4.72\times10^{-4}$    & -      \\
\hline
Ours                 & $1.25\times10^{-4}$               & $5.95\times10^{-2}$                & $3.11\times10^{-4}$    & $1.57\times10^{-2}$     \\
\hline
\end{tabular}
\end{center}
\end{table}

\subsection{Comparison of $Pe$ Values on the Two Datasets}

As discussed earlier, molecular transport within the ECS occurs through advection and diffusion, but the precise quantification of their relative contributions remains a challenge. Leveraging the data approximation capabilities of PINN, we can automatically estimate the diffusion coefficient and molecular velocity, allowing for a quantitative and accurate assessment of the importance of these transport forms. The P{\'e}clet number, $Pe$, offers a rough estimation of the relative significance of advection and diffusion in low permeability environments~\cite{huysmans2005review,nicholson2017brain}, defined as:
\begin{equation}
Pe=\frac{Lv}{D}\label{pe}
\end{equation}
where $L$ is the characteristic length, $v$ is the velocity of molecular and $D$ is the diffusion coefficient. Taking~\cite{nicholson2017brain} as a reference, we set $L$ as $100$ $\mu$m and calculate $Pe$ according to~(\ref{pe}) with the estimated $D$ and $v$ from the two datasets. The results of $Pe$ on the two datasets are presented in Table~\ref{comp_pe}.

For dataset-1, we find $Pe=47.60>1$, indicating that advection may dominate molecular transport within the ECS over the typical observation distance. This suggests that advection plays a significant role in the molecular transport, and diffusion can be negligible in this scenario. Similarly, for dataset-2, we have $Pe=5.05>1$, indicating that advection may dominate molecular transport within the ECS, and diffusion is not crucial in this context. The consistent results of $Pe$ calculated for both datasets using the proposed method demonstrate that advection is the primary mode of molecular transport in the analyzed region of the caudate nucleus.

\begin{table}[!t]
\renewcommand{\arraystretch}{1.5}
\caption{Comparison of $Pe$ values on the two datasets}
\label{comp_pe}
\begin{center}
\begin{tabular}{|c|c|c|}
\hline
{\bf Dataset}            & $\mathbf{Pe}$              & {\bf Diffusion or advection}             \\
\hline
Dataset-1          & 47.60              & Advection                          \\
\hline
Dataset-2          & 5.05               & Advection                          \\
\hline
\end{tabular}
\end{center}
\end{table}

\section{Conclusions and future work}
The nature of molecular transport within the brain ECS poses a challenging and unresolved problem. This paper addresses this issue by investigating it through the application of the advection-diffusion equation. The equation is effectively solved using a physics-informed neural network, enabling automatic computation of the diffusion coefficient and molecular velocity through the network optimization. Consequently, the specific mode of molecular transport within the ECS can the quantitatively analyzed and identified through the calculation of the P{\'e}clet number. Extensive numerical results, obtained from two datasets featuring MRI data acquired at different time points, illustrate the consistency and reliability of the proposed approach when compared to the existing reference. These findings suggest that the physics-informed neural network holds promise as a valuable tool for studying molecular transport within the ECS. In future research, we aim to investigate more advanced network structures and weight learning methods to enhance the performance of the PINNs for the presented problem. Furthermore, we plan to leverage advanced imaging tools, such as cryo-electron microscopy (cryo-EM), to acquire precise ECS microstructure information. This information will enable the development of more accurate mathematical models and efficient optimization algorithms, thereby advancing our understanding of molecular transport within the ECS.

To ensure reproducibility, the source code for this paper will be made publicly accessible online upon acceptance.

\ifCLASSOPTIONcaptionsoff
  \newpage
\fi



%


%

\bibliographystyle{./IEEEtran}
\bibliography{mybibfile}

\begin{thebibliography}{10}
\providecommand{\url}[1]{#1}
\csname url@samestyle\endcsname
\providecommand{\newblock}{\relax}
\providecommand{\bibinfo}[2]{#2}
\providecommand{\BIBentrySTDinterwordspacing}{\spaceskip=0pt\relax}
\providecommand{\BIBentryALTinterwordstretchfactor}{4}
\providecommand{\BIBentryALTinterwordspacing}{\spaceskip=\fontdimen2\font plus
\BIBentryALTinterwordstretchfactor\fontdimen3\font minus
  \fontdimen4\font\relax}
\providecommand{\BIBforeignlanguage}[2]{{%
\expandafter\ifx\csname l@#1\endcsname\relax
\typeout{** WARNING: IEEEtran.bst: No hyphenation pattern has been}%
\typeout{** loaded for the language `#1'. Using the pattern for}%
\typeout{** the default language instead.}%
\else
\language=\csname l@#1\endcsname
\fi
#2}}
\providecommand{\BIBdecl}{\relax}
\BIBdecl

\bibitem{sykova2008diffusion}
E.~Sykov{\'a} and C.~Nicholson, ``Diffusion in brain extracellular space,''
  \emph{Physiol. Rev.}, vol.~88, no.~4, pp. 1277--1340, 2008.

\bibitem{wang2021alteration}
R.~Wang, H.~Han, K.~Shi, I.~L. Alberts, A.~Rominger, C.~Yang, J.~Yan, D.~Cui,
  Y.~Peng, Q.~He, Y.~Gao, J.~Lian, S.~Yang, H.~Liu, J.~Yang, C.~Wong, X.~Wei,
  F.~Yin, Y.~Jia, H.~Tong, B.~Liu, and J.~Lei, ``The alteration of brain
  interstitial fluid drainage with myelination development,'' \emph{Aging
  Dis.}, vol.~12, no.~7, p. 1729, 2021.

\bibitem{iliff2012paravascular}
J.~J. Iliff, M.~Wang, Y.~Liao, B.~A. Plogg, W.~Peng, G.~A. Gundersen,
  H.~Benveniste, G.~E. Vates, R.~Deane, S.~A. Goldman \emph{et~al.}, ``A
  paravascular pathway facilitates csf flow through the brain parenchyma and
  the clearance of interstitial solutes, including amyloid $\beta$,''
  \emph{Sci. Transl. Med.}, vol.~4, no. 147, p. 147ra111, 2012.

\bibitem{li2020mechanism}
Y.~Li, H.~Han, K.~Shi, D.~Cui, J.~Yang, I.~L. Alberts, L.~Yuan, G.~Zhao,
  R.~Wang, X.~Cai, and Z.~Teng, ``The mechanism of downregulated interstitial
  fluid drainage following neuronal excitation,'' \emph{Aging Dis.}, vol.~11,
  no.~6, p. 1407, 2020.

\bibitem{lu2014integrated}
Y.~Lu, K.~Yang, K.~Zhou, B.~Pang, G.~Wang, Y.~Ding, Q.~Zhang, H.~Han, J.~Tian,
  C.~Li, and Q.~Ren, ``An integrated quad-modality molecular imaging system for
  small animals,'' \emph{J. Nucl. Med.}, vol.~55, no.~8, pp. 1375--1379, 2014.

\bibitem{wiig2012interstitial}
H.~Wiig and M.~A. Swartz, ``Interstitial fluid and lymph formation and
  transport: physiological regulation and roles in inflammation and cancer,''
  \emph{Physiol. Rev.}, vol.~92, no.~3, pp. 1005--1060, 2012.

\bibitem{gu2022new}
Z.~Gu, H.~Chen, H.~Zhao, W.~Yang, Y.~Song, X.~Li, Y.~Wang, D.~Du, H.~Liao,
  W.~Pan \emph{et~al.}, ``New insight into brain disease therapy:
  nanomedicines-crossing blood--brain barrier and extracellular space for drug
  delivery,'' \emph{Expert Opin. Drug Deliv.}, vol.~19, no.~12, pp. 1618--1635,
  2022.

\bibitem{zhou2013protective}
N.~Zhou, T.~Xu, Y.~Bai, S.~Prativa, J.-Z. Xu, K.~Li, H.-B. Han, and J.-H. Yan,
  ``Protective effects of urinary trypsin inhibitor on vascular permeability
  following subarachnoid hemorrhage in a rat model,'' \emph{CNS Neurosci.
  Ther.}, vol.~19, no.~9, pp. 659--666, 2013.

\bibitem{ferguson2007convection}
S.~D. Ferguson, K.~Foster, and B.~Yamini, ``Convection-enhanced delivery for
  treatment of brain tumors,'' \emph{Expert Rev. Anticancer Ther}, vol.~7, no.
  12 Supp1, pp. S79--S85, 2007.

\bibitem{han2011simple}
H.~Han, Z.~Xia, H.~Chen, C.~Hou, and W.~Li, ``Simple diffusion delivery via
  brain interstitial route for the treatment of cerebral ischemia,'' \emph{Sci.
  China-Life Sci.}, vol.~54, no.~3, pp. 235--239, 2011.

\bibitem{gao2022early}
Y.~Gao, H.~Han, J.~Du, Q.~He, Y.~Jia, J.~Yan, H.~Dai, B.~Cui, J.~Yang, X.~Wei
  \emph{et~al.}, ``Early changes to the extracellular space in the hippocampus
  under simulated microgravity conditions,'' \emph{Sci. China-Life Sci.},
  vol.~65, no.~3, pp. 604--617, 2022.

\bibitem{han2014novel}
H.~Han, C.~Shi, Y.~Fu, L.~Zuo, K.~Lee, Q.~He, and H.~Han, ``A novel mri
  tracer-based method for measuring water diffusion in the extracellular space
  of the rat brain,'' \emph{IEEE J. Biomed. Health Inform.}, vol.~18, no.~3,
  pp. 978--983, 2014.

\bibitem{vendel2019need}
E.~Vendel, V.~Rottsch{\"a}fer, and E.~C. de~Lange, ``The need for mathematical
  modelling of spatial drug distribution within the brain,'' \emph{Fluids
  Barriers CNS}, vol.~16, p.~12, 2019.

\bibitem{iliff2019glymphatic}
J.~Iliff and M.~Simon, ``Crosstalk proposal: The glymphatic system supports
  convective exchange of cerebrospinal fluid and brain interstitial fluid that
  is mediated by perivascular aquaporin-4,'' \emph{J. Physiol.-London}, vol.
  597, no.~17, pp. 4417--4419, 2019.

\bibitem{smith2017test}
A.~J. Smith, X.~Yao, J.~A. Dix, B.-J. Jin, and A.~S. Verkman, ``Test of
  the'glymphatic'hypothesis demonstrates diffusive and aquaporin-4-independent
  solute transport in rodent brain parenchyma,'' \emph{eLife}, vol.~6, p.
  e27679, 2017.

\bibitem{iliff2014impairment}
J.~J. Iliff, M.~J. Chen, B.~A. Plog, D.~M. Zeppenfeld, M.~Soltero, L.~Yang,
  I.~Singh, R.~Deane, and M.~Nedergaard, ``Impairment of glymphatic pathway
  function promotes tau pathology after traumatic brain injury,'' \emph{J.
  Neurosci.}, vol.~34, no.~49, pp. 16\,180--16\,193, 2014.

\bibitem{wang2019drainage}
A.~Wang, R.~Wang, D.~Cui, X.~Huang, L.~Yuan, H.~Liu, Y.~Fu, L.~Liang, W.~Wang,
  Q.~He \emph{et~al.}, ``The drainage of interstitial fluid in the deep brain
  is controlled by the integrity of myelination,'' \emph{Aging Dis.}, vol.~10,
  no.~5, pp. 937--948, 2019.

\bibitem{elkin2018glymphvis}
R.~Elkin, S.~Nadeem, E.~Haber, K.~Steklova, H.~Lee, H.~Benveniste, and
  A.~Tannenbaum, ``Glymphvis: visualizing glymphatic transport pathways using
  regularized optimal transport,'' in \emph{21st Medical Image Computing and
  Computer Assisted Intervention (MICCAI 2018), Proceedings, Part I}, Spain,
  Sep. 2018, pp. 844--852.

\bibitem{koundal2020optimal}
S.~Koundal, R.~Elkin, S.~Nadeem, Y.~Xue, S.~Constantinou, S.~Sanggaard, X.~Liu,
  B.~Monte, F.~Xu, W.~Van~Nostrand \emph{et~al.}, ``Optimal mass transport with
  lagrangian workflow reveals advective and diffusion driven solute transport
  in the glymphatic system,'' \emph{Sci Rep}, vol.~10, no.~1, p. 1990, 2020.

\bibitem{liu2007stability}
F.~Liu, P.~Zhuang, V.~Anh, I.~Turner, and K.~Burrage, ``Stability and
  convergence of the difference methods for the space--time fractional
  advection--diffusion equation,'' \emph{Appl. Math. Comput.}, vol. 191, no.~1,
  pp. 12--20, 2007.

\bibitem{wang2019stimulation}
W.~Wang, Q.~He, J.~Hou, D.~Chui, M.~Gao, A.~Wang, H.~Han, and H.~Liu,
  ``Stimulation modeling on three-dimensional anisotropic diffusion of mri
  tracer in the brain interstitial space,'' \emph{Front. Neuroinformatics},
  vol.~13, p.~6, 2019.

\bibitem{monge1781memoire}
G.~Monge, \emph{M{\'e}moire sur la th{\'e}orie des d{\'e}blais et des
  remblais}.\hskip 1em plus 0.5em minus 0.4em\relax De l'Imprimerie Royale,
  1781.

\bibitem{benamou2000computational}
J.-D. Benamou and Y.~Brenier, ``A computational fluid mechanics solution to the
  monge-kantorovich mass transfer problem,'' \emph{Numer. Math.}, vol.~84,
  no.~3, pp. 375--393, 2000.

\bibitem{fan2020effect}
C.~Fan, F.~Tian, X.~Zhao, Y.~Sun, X.~Yang, H.~Han, and X.~Pu, ``The effect of
  thymoquinone on the characteristics of the brain extracellular space in
  transient middle cerebral artery occlusion rats,'' \emph{Biol. Pharm. Bull.},
  vol.~43, no.~9, pp. 1306--1314, 2020.

\bibitem{ratner2017cerebrospinal}
V.~Ratner, Y.~Gao, H.~Lee, R.~Elkin, M.~Nedergaard, H.~Benveniste, and
  A.~Tannenbaum, ``Cerebrospinal and interstitial fluid transport via the
  glymphatic pathway modeled by optimal mass transport,'' \emph{Neuroimage},
  vol. 152, pp. 530--537, 2017.

\bibitem{ratner2015optimal}
V.~Ratner, L.~Zhu, I.~Kolesov, M.~Nedergaard, H.~Benveniste, and A.~Tannenbaum,
  ``Optimal-mass-transfer-based estimation of glymphatic transport in living
  brain,'' in \emph{Medical Imaging 2015: Image Processing}, vol. 9413.\hskip
  1em plus 0.5em minus 0.4em\relax SPIE, 2015, p. 94131J.

\bibitem{chen2023visualizing}
X.~Chen, A.~P. Tran, R.~Elkin, H.~Benveniste, and A.~R. Tannenbaum,
  ``Visualizing fluid flows via regularized optimal mass transport with
  applications to neuroscience,'' \emph{J. Sci. Comput.}, vol.~97, no.~2,
  p.~26, 2023.

\bibitem{chen2022cerebral}
X.~Chen, X.~Liu, S.~Koundal, R.~Elkin, X.~Zhu, B.~Monte, F.~Xu, F.~Dai,
  M.~Pedram, H.~Lee \emph{et~al.}, ``Cerebral amyloid angiopathy is associated
  with glymphatic transport reduction and time-delayed solute drainage along
  the neck arteries,'' \emph{Nature Aging}, vol.~2, no.~3, pp. 214--223, 2022.

\bibitem{mardal2022mathematical}
K.-A. Mardal, M.~E. Rognes, T.~B. Thompson, and L.~M. Valnes,
  \emph{Mathematical Modeling of the Human Brain: From Magnetic Resonance
  Images to Finite Element Simulation}.\hskip 1em plus 0.5em minus 0.4em\relax
  Springer Nature, 2022.

\bibitem{valnes2020apparent}
L.~M. Valnes, S.~K. Mitusch, G.~Ringstad, P.~K. Eide, S.~W. Funke, and K.-A.
  Mardal, ``Apparent diffusion coefficient estimates based on 24 hours tracer
  movement support glymphatic transport in human cerebral cortex,'' \emph{Sci
  Rep}, vol.~10, no.~1, p. 9176, 2020.

\bibitem{li2018neural}
H.~Li, H.~Zhao, and H.~Li, ``Neural-response-based extreme learning machine for
  image classification,'' \emph{IEEE Trans. Neural Netw. Learn. Syst.},
  vol.~30, no.~2, pp. 539--552, 2019.

\bibitem{raissi2019physics}
M.~Raissi, P.~Perdikaris, and G.~E. Karniadakis, ``Physics-informed neural
  networks: A deep learning framework for solving forward and inverse problems
  involving nonlinear partial differential equations,'' \emph{J. Comput.
  Phys.}, vol. 378, pp. 686--707, 2019.

\bibitem{raissi2018hidden}
M.~Raissi and G.~E. Karniadakis, ``Hidden physics models: Machine learning of
  nonlinear partial differential equations,'' \emph{J. Comput. Phys.}, vol.
  357, pp. 125--141, 2018.

\bibitem{yang2023physics}
Q.~Yang, Z.~Wang, K.~Guo, C.~Cai, and X.~Qu, ``Physics-driven synthetic data
  learning for biomedical magnetic resonance: The imaging physics-based data
  synthesis paradigm for artificial intelligence,'' \emph{IEEE Signal Process.
  Mag.}, vol.~40, no.~2, pp. 129--140, 2023.

\bibitem{cai2023bloch}
Q.~Cai, L.~Zhu, J.~Zhou, C.~Qian, D.~Guo, and X.~Qu, ``Bloch equation enables
  physics-informed neural network in parametric magnetic resonance imaging,''
  \emph{arXiv preprint arXiv:2309.11763}, 2023.

\bibitem{kapoor2023physics}
T.~Kapoor, H.~Wang, A.~Nunez, and R.~Dollevoet, ``Physics-informed neural
  networks for solving forward and inverse problems in complex beam systems,''
  \emph{IEEE Trans. Neural Netw. Learn. Syst.}, 2023, doi:
  10.1109/TNNLS.2023.3310585.

\bibitem{cui2023knowledge}
Z.~Cui, T.~Gao, K.~Talamadupula, and Q.~Ji, ``Knowledge-augmented deep learning
  and its applications: A survey,'' \emph{IEEE Trans. Neural Netw. Learn.
  Syst.}, 2023, doi: 10.1109/TNNLS.2023.3338619.

\bibitem{zapf2022investigating}
B.~Zapf, J.~Haubner, M.~Kuchta, G.~Ringstad, P.~K. Eide, and K.-A. Mardal,
  ``Investigating molecular transport in the human brain from mri with
  physics-informed neural networks,'' \emph{Sci Rep}, vol.~12, no.~1, p. 15475,
  2022.

\bibitem{van2022physics}
R.~L. van Herten, A.~Chiribiri, M.~Breeuwer, M.~Veta, and C.~M. Scannell,
  ``Physics-informed neural networks for myocardial perfusion mri
  quantification,'' \emph{Med. Image Anal.}, vol.~78, p. 102399, 2022.

\bibitem{sarabian2022physics}
M.~Sarabian, H.~Babaee, and K.~Laksari, ``Physics-informed neural networks for
  brain hemodynamic predictions using medical imaging,'' \emph{IEEE Trans. Med.
  Imaging}, vol.~41, no.~9, pp. 2285--2303, 2022.

\bibitem{oszkinat2022uncertainty}
C.~Oszkinat, S.~E. Luczak, and I.~Rosen, ``Uncertainty quantification in
  estimating blood alcohol concentration from transdermal alcohol level with
  physics-informed neural networks,'' \emph{IEEE Trans. Neural Netw. Learn.
  Syst.}, vol.~34, no.~10, pp. 8094--8101, 2023.

\bibitem{zhang2023physics}
X.~Zhang, B.~Mao, Y.~Che, J.~Kang, M.~Luo, A.~Qiao, Y.~Liu, H.~Anzai, M.~Ohta,
  Y.~Guo \emph{et~al.}, ``Physics-informed neural networks (pinns) for 4d
  hemodynamics prediction: An investigation of optimal framework based on
  vascular morphology,'' \emph{Comput. Biol. Med.}, vol. 164, p. 107287, 2023.

\bibitem{moser2023modeling}
P.~Moser, W.~Fenz, S.~Thumfart, I.~Ganitzer, and M.~Giretzlehner, ``Modeling of
  3d blood flows with physics-informed neural networks: Comparison of network
  architectures,'' \emph{Fluids}, vol.~8, no.~2, p.~46, 2023.

\bibitem{kingma2014adam}
D.~P. Kingma and J.~Ba, ``Adam: A method for stochastic optimization,''
  \emph{arXiv preprint arXiv:1412.6980}, 2014.

\bibitem{paszke2019pytorch}
A.~Paszke, S.~Gross, F.~Massa, A.~Lerer, J.~Bradbury, G.~Chanan, T.~Killeen,
  Z.~Lin, N.~Gimelshein, L.~Antiga \emph{et~al.}, ``Pytorch: An imperative
  style, high-performance deep learning library,'' \emph{Advances in neural
  information processing systems}, vol.~32, 2019.

\bibitem{huysmans2005review}
M.~Huysmans and A.~Dassargues, ``Review of the use of p{\'e}clet numbers to
  determine the relative importance of advection and diffusion in low
  permeability environments,'' \emph{Hydrogeol. J.}, vol.~13, no. 5--6, pp.
  895--904, 2005.

\bibitem{nicholson2017brain}
C.~Nicholson and S.~Hrab{\v{e}}tov{\'a}, ``Brain extracellular space: the final
  frontier of neuroscience,'' \emph{Biophys. J.}, vol. 113, no.~10, pp.
  2133--2142, 2017.

\end{thebibliography}

%

%
%
%




\end{document}